\documentclass[10pt,twocolumn,letterpaper]{article}

\usepackage{cvpr}              % To produce the CAMERA-READY version
\usepackage{multirow}
\usepackage{booktabs}       % professional-quality tables
\usepackage{amsfonts}       % blackboard math symbols
\usepackage{nicefrac}       % compact symbols for 1/2, etc.
\usepackage{microtype}      % microtypography
\usepackage{comment}
\usepackage{amsmath}
\usepackage{adjustbox}
\usepackage{caption}
\usepackage{bm}
\usepackage{enumitem}
\usepackage{float}
\usepackage{wrapfig}
\usepackage{subcaption}
\usepackage[utf8]{inputenc} % allow utf-8 input
\usepackage[T1]{fontenc}    % use 8-bit T1 fonts
\usepackage[accsupp]{axessibility} 

\usepackage[table]{xcolor} 

\definecolor{cvprblue}{rgb}{0.21,0.49,0.74}
\usepackage[pagebackref,breaklinks,colorlinks,allcolors=cvprblue]{hyperref}

\def\paperID{} % *** Enter the Paper ID here
\def\confName{CVPR}
\def\confYear{2026}

\title{UCAN: Unified Convolutional Attention Network for Expansive Receptive Fields in Lightweight Super-Resolution}

\author{
    Cao Thien Tan$^{1,2,3}$ \quad 
    Phan Thi Thu Trang $^{3,5}$ \quad 
    Do Nghiem Duc$^6$ \quad 
    Ho Ngoc Anh$^5$ \\ 
    Hanyang Zhuang$^{4,}$\footnotemark[1] \quad 
    Nguyen Duc Dung $^{2,}$\footnotemark[1]\\
    \\
    $^1$Ho Chi Minh City Open University \quad 
    $^2$AI Tech Lab, Ho Chi Minh City University of Technology\\ 
    $^3$Code Mely AI Research Team\quad 
    $^4$Global College, Shanghai Jiao Tong University\\ 
    $^5$Ha Noi University of Science and Technology \quad  
    $^6$University of Manitoba\\
    {\tt\small caothientan2001@gmail.com, zhuanghany11@sjtu.edu.cn, nddung@hcmut.edu.vn}\\
    \small Code: \url{https://github.com/hokiyoshi/UCAN}
} 

\begin{document}

\maketitle

% Force the footnote text to appear on the first page
\renewcommand{\thefootnote}{\fnsymbol{footnote}}
\footnotetext[1]{Corresponding author}
\renewcommand{\thefootnote}{\arabic{footnote}} % Reset to numbers for the rest of the paper

\begin{abstract}
Hybrid CNN-Transformer architectures achieve strong results in image super-resolution, but scaling attention windows or convolution kernels significantly increases computational cost, limiting deployment on resource-constrained devices. We present UCAN, a lightweight network that unifies convolution and attention to expand the effective receptive field efficiently. UCAN combines window-based spatial attention with a Hedgehog Attention mechanism to model both local texture and long-range dependencies, and introduces a distillation-based large-kernel module to preserve high-frequency structure without heavy computation. In addition, we employ cross-layer parameter sharing to further reduce complexity. On Manga109 ($4\times$), UCAN-L achieves 31.63 dB PSNR with only 48.4G MACs, surpassing recent lightweight models. On BSDS100, UCAN attains 27.79 dB, outperforming methods with significantly larger models. Extensive experiments show that UCAN achieves a superior trade-off between accuracy, efficiency, and scalability, making it well-suited for practical high-resolution image restoration. 
\end{abstract}

\section{Introduction}
\vspace{-2mm}
Single-image super-resolution (SR) aims to reconstruct high resolution (HR) images from degraded low resolution (LR) inputs, representing a long-standing yet continually evolving challenge in low-level vision. This task has garnered sustained attention in the computer vision community, driven by its broad spectrum of practical applications and the increasing demand for both accuracy and efficiency in SR solutions. Recent advancements have seen the emergence of lightweight SR models, with growing emphasis on enhancing performance by expanding the effective receptive field. While transformer-based architectures have demonstrated strong capability in this regard, recent research investigates alternative strategies that extend the receptive field through new architectures or techniques with lower computational complexity and fewer parameters.
% Add citation

% More recently, the selective state-space model (Mamba) \cite{gu2023mamba}  has been explored as an alternative architecture for image restoration tasks \cite{shi2024vmambair, guo2025mambair}. However, previous research \cite{guo2024mambairv2} has demonstrated that the effectiveness of Mamba is constrained by the decay of long-range memory.

\begin{figure}[t]
\centering
\includegraphics[width=1\columnwidth]{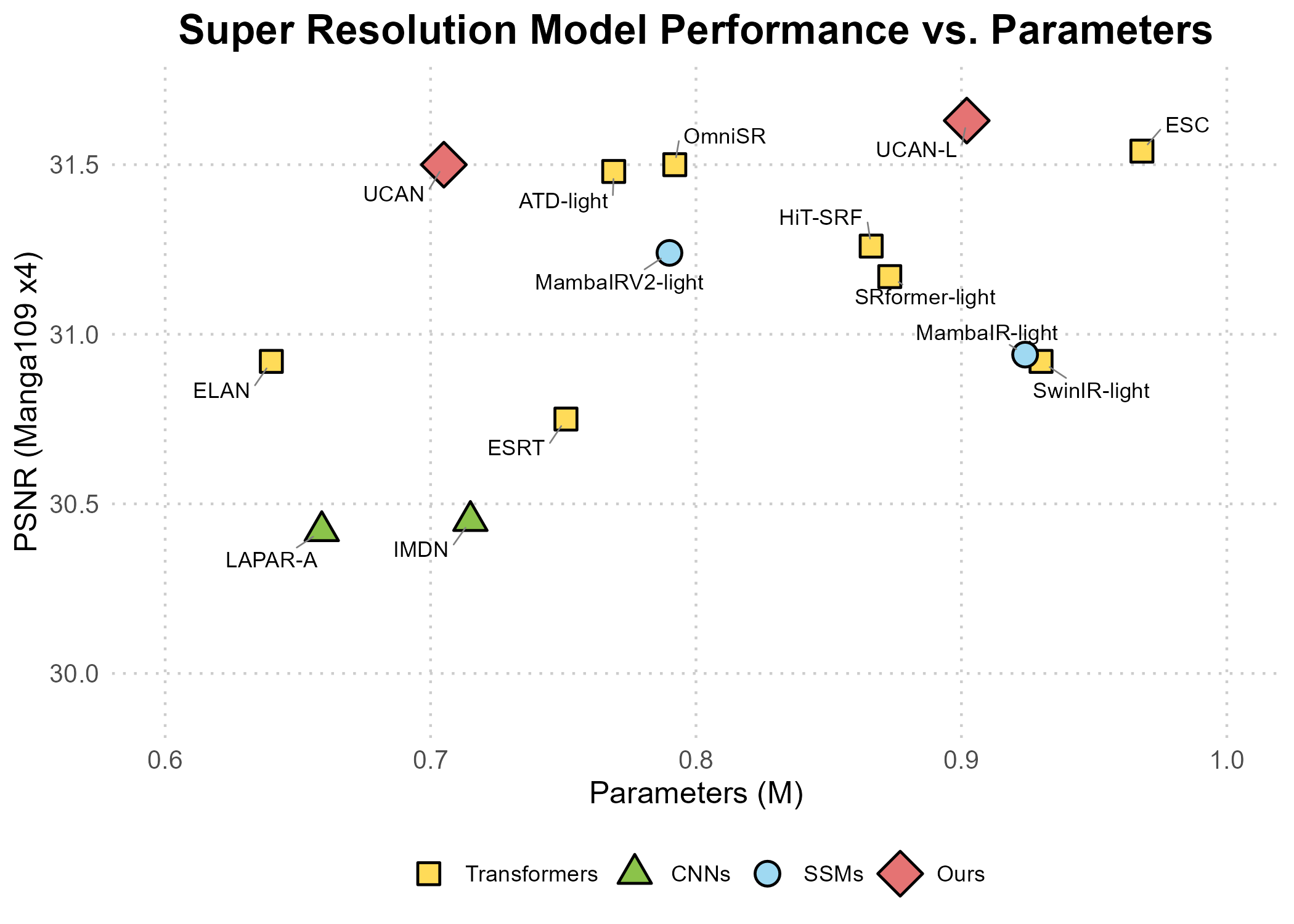} % Reduce the figure size so that it is slightly narrower than the column. Don't use precise values for figure width.This setup will avoid overfull boxes.
\caption{Performance comparison of PSNR versus model parameters on the Manga109 ($\times$4) dataset. Our method is evaluated alongside state-of-the-art super-resolution approaches. Green triangles represent CNN-based methods, yellow squares denote Transformer-based methods, blue circles indicate SSM-based methods, and red diamonds show our proposed approach.}
\label{fig:Intro}
\end{figure}

Recent advances in lightweight super-resolution primarily focus on enlarging the effective receptive field to capture broader contextual information and recover high-frequency details. While this expansion is essential, existing approaches remain limited. Convolution-based architectures effectively model local dependencies, yet many efforts to introduce global attention such as Grid Attention \cite{wang2023omni}, Mamba \cite{shi2024vmambair, guo2024mambairv2}, and Adaptive Token Dictionary \cite{zhang2024transcending} still struggle with inefficiency in capturing global context. Linear attention \cite{ai2025breaking} offers a promising alternative, but its potential is constrained by limited feature diversity, which weaken representational capacity. Moreover, the trade-off between efficiency and performance persists; although recent models achieve lower parameter counts, they often sacrifice representational richness. Joint attention mechanisms and distillation strategies \cite{park2025efficient} further homogenize feature maps across layers and channels, reducing feature diversity and ultimately hindering reconstruction quality.

% !!Caution: Fix the performance to my own experiment
We introduce the Unified Convolutional Attention Network (UCAN), a lightweight super-resolution architecture that expands the effective receptive field while preserving rich feature representation. The High Performance Block incorporates Flash Attention to avoid computing the full attention matrix, allowing an efficient $32\times32$ window and achieving substantially lower latency compared to conventional attention. To capture fine local structures, long-range dependencies, and channel relationships, we design a Hybrid Attention mechanism where windowed attention efficiently models local patterns, and the Dual Fusion Layer, an optimized implementation of Hedgehog Attention, addresses the rank deficiency typically observed in linear and channel attention. In addition, parameter sharing is employed to reduce computational cost while maintaining strong performance. A Large Kernel Distillation module further enhances the network’s ability to reproduce complex textures and hierarchical structures with minimal parameter overhead. Across standard benchmarks, UCAN delivers higher visual fidelity while preserving a favorable balance between latency and computational efficiency.

Building upon the aforementioned components, \textbf{UCAN} demonstrates a strong balance between lightweight design and expressive capability. As shown in Fig.~\ref{fig:Intro}, UCAN achieves competitive performance with a favorable trade-off between parameter count and reconstruction accuracy. On standard benchmarks for $2\times$ upscaling, UCAN surpasses the OmniSR model by \textbf{0.12\,dB} on Urban100 and \textbf{0.17\,dB} on Set5, while using \textbf{11\% fewer parameters} and \textbf{24\% fewer FLOPs}. For $4\times$ upscaling on Manga109, the larger variant UCAN-L outperforms MambaIRV2 by \textbf{0.39\,dB}, achieving up to \textbf{36\% lower computational cost}. These consistent improvements highlight the effectiveness and efficiency of our unified convolutional attention framework in achieving high-fidelity image reconstruction.

In summary, our contributions are given as follows:

\begin{itemize}
    % \item We propose a semi-sharing mechanism that reduces attention computation while preserving representational diversity. 
    
    % \item We integrate Flash Attention with $32\times32$ windows and linear attention via Dual Fusion Layers to balance local detail capture and global context modeling. This hybrid approach extends effective receptive fields while maintaining computational efficiency.
    
    % \item We introduce Large Kernel Distillation, which transfers spatial knowledge from dilated convolutions to attention modules through feature partitioning, achieving large receptive fields without quadratic complexity scaling.

    \item We introduce Hedgehog Attention, a novel attention mechanism that enhances feature diversity in linear attention by enhancing the rank of features. This design allows the model to capture richer and more expressive features.
    \item We propose UCAN, a unified and efficient model that captures both local and global dependencies across multiple receptive fields. UCAN employs Flash Attention for large-window attention to improve computational efficiency, Hedgehog Attention to model global information with strong robustness under limited resources, and multi-kernel convolution to extract essential local features.

    %To the best of our knowledge, this is the first work to demonstrate the effectiveness of integrating these four complementary mechanisms in single image super resolution tasks.
    \item We design a semi-sharing and distillation architecture that balances parameter sharing with task-specific specialization. Experimental results show that this architecture allows UCAN to achieve strong performance while substantially reducing the number of parameters and computational cost.   
\end{itemize}

\section{Related work}

% Early deep learning-based image super-resolution methods \cite{dong2014learning, dong2015compression} primarily relied on stacked convolutions for local feature extraction, which often resulted in over-parameterization. To address this issue, numerous studies \cite{kim2016deeply, tai2017image, hui2018fast} explored weight sharing strategies across multiple layers, though many suffered from performance degradation due to input-independent weight sharing. Recent advancements have enhanced CNN representational capacity through architectural innovations. EDSR \cite{lim2017enhanced} employed residual connections for very deep networks, RDN \cite{zhang2018residual} utilized dense connections for improved feature reuse, and RCAN \cite{zhang2018image} introduced channel attention for salient feature selection. Despite these improvements, the convolution operator inherently restricts the receptive field to local kernels, preventing interaction between distant pixels and motivating the exploration of architectures with global dependency modeling capabilities.

\textbf{Convolutional Methods.} Early deep learning-based SR methods \cite{dong2014learning, dong2015compression} employed stacked convolutions for local feature extraction but suffered from over-parameterization. Weight-sharing strategies \cite{kim2016deeply, tai2017image, hui2018fast} reduced complexity, yet input-independent sharing often degraded performance. Later architectures improved CNN expressiveness: EDSR \cite{lim2017enhanced} with deep residual blocks, RDN \cite{zhang2018residual} with dense feature reuse, and RCAN \cite{zhang2018image} with channel attention. Despite these advances, convolutions remain constrained to local receptive fields, motivating the exploration of architectures that model long-range dependencies.
\paragraph{Comparative Analysis of Attention Methods.}
\vspace{-1em}
Vision Transformers have recently gained traction in image super-resolution following their success in general computer vision tasks \cite{dosovitskiy2020image, ranftl2021vision}. Representative methods include SwinIR \cite{liu2021swin}, which adapts the shifted-window strategy for image restoration, ESRT \cite{lu2022transformer}, which combines lightweight CNNs with transformer blocks for efficiency, and ELAN \cite{zhang2022efficient}, which reduces cost through shared self-attention and unified query-key parameters \cite{zamir2022restormer}. 

Despite these advances, contemporary transformer architectures still suffer from limited effective receptive fields, restricting global context modeling and computational efficiency. To address this, Omni-SR \cite{wang2023omni} introduced grid attention to enlarge spatial observation, while MambaIRv2 \cite{guo2024mambairv2} combined Swin Transformer with State Space Models to enhance whole-image information capture. Building on these ideas, our approach integrates Swin Transformer with an improved Linear Transformer, achieving superior performance by extending the receptive field while remaining efficient.
\paragraph{Large Receptive Field Attention Mechanisms.}
\vspace{-1em}
The computational cost of Transformers has driven research into CNN-based alternatives that use large kernels and pixel-level attention to approximate Transformer performance \cite{xie2023large,wu2024transforming,sun2022shufflemixer}. However, these methods often fail to capture self-attention's rich representations or incur memory overhead from complex architectures like multi-directional scanning.
Recent work has addressed CNN kernel expansion through CUDA optimization \cite{lau2024large}, asymmetric convolutions \cite{szegedy2016rethinking}, dilated convolutions \cite{lau2024large}, and distillation techniques \cite{lee2025emulating}. Building on these advances, we propose Large Kernel Distillation, which combines dilated convolutions with knowledge distillation to achieve large receptive fields while maintaining computational efficiency.

\begin{figure}[t]
\centering
\includegraphics[width=1.01\columnwidth]{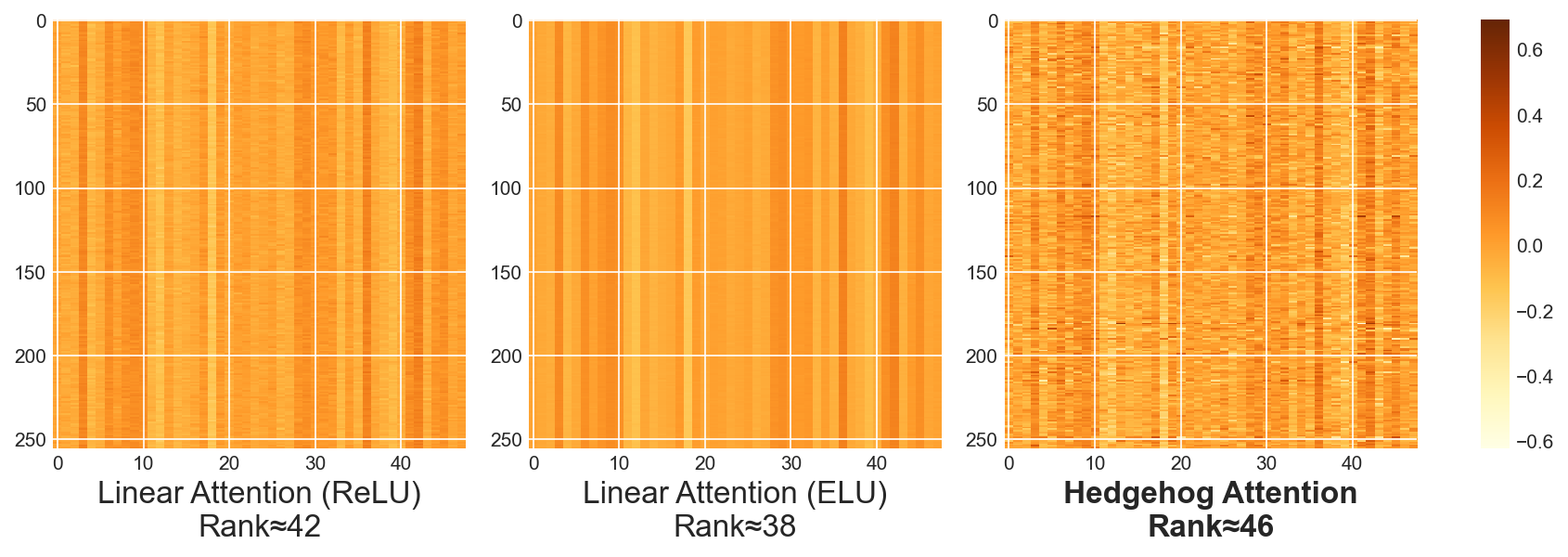} 
\caption{Comparison of feature maps output by Linear Attention (Using ReLU, ELU feature maps) and Hedgehog Attention. All experiments are conducted based on an image with $N$ = 256 and $d$ = 48. The full rank of matrices in the figure is 64. This figure shows that our model outperforms previous methods in diversity.}
\label{fig:ranking_output}
\end{figure}

\section{Proposed Method}
In this section, we begin with an analysis of Linear Transformers and Softmax Attention, and then provide a detailed description of the proposed network.

\begin{figure*}[h]
\centering
\includegraphics[width=2.7\columnwidth]{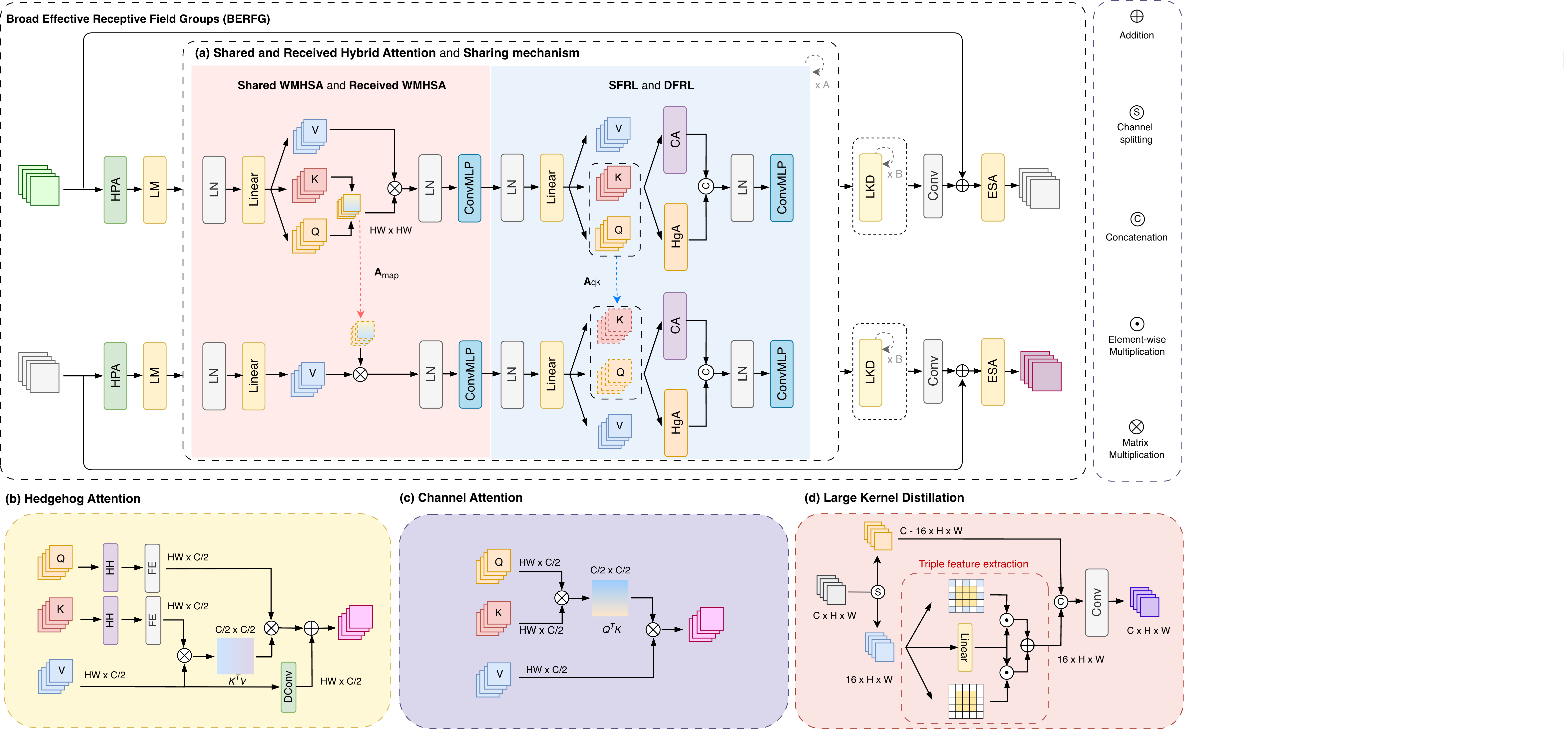} 
\vspace{0mm}
\caption{Detailed architecture of (a) Shared and Received Hybrid Attention (SHA and RHA) and (b) Large Kernel Distillation (LKD). LKD contains a Triple Feature Extraction block with three branches, which are the Large Kernel Spatial Branch, the Channel Branch, and the Small Kernel Spatial Branch. SHA and RHA employ Shared and Received Window Multi Head Self Attention (Shared  WMHSA and Received  WMHSA) to capture local information, and include a Shared Dual Fusion Layer (SDFL) and a Dual Fusion Receiver Layer (DFRL) to aggregate global context. The Dual Fusion Layer comprises two sub branches, which are Hedgehog Attention (HgA) and Channel Attention (CA).}
\vspace{0mm}
\label{figDetailarch}
\end{figure*}

\subsection{Bridging Softmax and Linear Attention}
\paragraph{Connection Analysis.} Let \(\boldsymbol{X}\in\mathbb{R}^{HW\times C}\) and \(\boldsymbol{Q}=\boldsymbol{X}\boldsymbol{W}_Q,\; \boldsymbol{K}=\boldsymbol{X}\boldsymbol{W}_K,\; \boldsymbol{V}=\boldsymbol{X}\boldsymbol{W}_V\) with \(\boldsymbol{W}_Q,\boldsymbol{W}_K,\boldsymbol{W}_V\in\mathbb{R}^{C\times d}\). Denote rows by \(\boldsymbol{q}_i,\boldsymbol{k}_i,\boldsymbol{v}_i\in\mathbb{R}^d\). Standard softmax attention is described as
\begin{equation}
\begin{aligned}
\boldsymbol{s}_i &= \left[ \frac{\exp(\boldsymbol{q}_i^\top \boldsymbol{k}_1)} {\sum_{j=1}^N \exp(\boldsymbol{q}_i^\top \boldsymbol{k}_j)}, \dots, \frac{\exp(\boldsymbol{q}_i^\top \boldsymbol{k}_{HW})} {\sum_{j=1}^N \exp(\boldsymbol{q}_i^\top \boldsymbol{k}_j)} \right]^\top,\\
\boldsymbol{o}_i^{\mathrm{S}} &= \boldsymbol{s}_i^\top \boldsymbol{V}.
\end{aligned}
\end{equation}

However, linear attention substitutes \(\exp(\boldsymbol{q}_i^\top \boldsymbol{k}_j)\) by \(\boldsymbol{\phi}(\boldsymbol{q}_i)^\top \boldsymbol{\phi}(\boldsymbol{k}_j)\)  for a feature map \(\boldsymbol{\phi} \in \mathbb{R}^d\to\mathbb{R}^r\)
\begin{equation}
\boldsymbol{o}_i^{\mathrm{L}} = \frac{\sum_{j=1}^N (\phi(\boldsymbol{q}_i)^\top\phi(\boldsymbol{k}_j))\,\boldsymbol{v}_j}{\sum_{j=1}^N \phi(\boldsymbol{q}_i)^\top\phi(\boldsymbol{k}_j)} = \frac{\phi(\boldsymbol{q}_i)^\top\!\big(\sum_{j=1}^N \phi(\boldsymbol{k}_j)\,\boldsymbol{v}_j^\top\big)}{\phi(\boldsymbol{q}_i)^\top\!\big(\sum_{j=1}^N \phi(\boldsymbol{k}_j)\big)}.
\end{equation}

The two aggregated terms \(\sum_j \boldsymbol{\phi}(\boldsymbol{k}_j)\,\boldsymbol{v}_j^\top\) and \(\sum_j \boldsymbol{\phi}(\boldsymbol{k}_j)\) are computable in \(\mathcal{O}(N)\) and reused for all queries, reducing per layer complexity from \(\mathcal{O}(N^2)\) to \(\mathcal{O}(N)\).

\paragraph{Breaking the low rank bottleneck in linear attention.} 
\vspace{-1em}
Previous work restores the rank of the output matrix by adding depthwise convolutions or auxiliary high rank multipliers, but this does not remove the intrinsic low rank tendency of linear attention. Rank collapse forces the model to rely on a few dependent directions and reduces representational diversity.

Approximating softmax in linear time requires a feature map \(\phi\) that reproduces softmax selectivity. Simple choices such as ReLU or ELU\(+1\) fail since ReLU discards negative contributions that can combine to form important positive attention under softmax, while ELU\(+1\) yields extreme relative variations that encourage rank collapse. A detailed analysis appears in the supplementary material.

We use the Hedgehog Feature Map (HFM) \cite{zhang2024hedgehog}. HFM concatenates \(m\) symmetric exponential feature pairs
\begin{equation}
\begin{aligned}
\boldsymbol{\phi}_{\mathrm{H}}(\boldsymbol{X}) = [\, \exp(\boldsymbol{W}^\top \boldsymbol{X} + \boldsymbol{b}_1), \dots, \exp(\boldsymbol{W}^\top \boldsymbol{X} + \boldsymbol{b}_m),\\
\exp(-\boldsymbol{W}^\top \boldsymbol{X} - \boldsymbol{b}_1), \dots, \exp(-\boldsymbol{W}^\top \boldsymbol{X} - \boldsymbol{b}_m)\,],
\end{aligned}
\end{equation}
where \(\boldsymbol{W}\in\mathbb{R}^{C\times C}\) is a shared projection and \(\{\boldsymbol{b}_i\}_{i=1}^m\) are learnable biases.

The HFM presents distinct advantages stemming from its core design. Its simple, trainable MLP style construction affords greater flexibility than rigid, hand designed feature maps. By promoting low entropy, spike like attention distributions, HFM also maintains fine grained, content dependent lookups while simultaneously increasing diversity. Moreover, its symmetric pairing mechanism preserves information from both positive and negative pre projection directions, thereby mitigating the information loss often caused by rectification. Consequently, HFM is able to suppress rank collapse directly, eliminating the need for external rank recovery modules. Our experimental results in Fig.~\ref{fig:ranking_output} show that the Linear Attention model applying HFM recovers a rank up to 46, exceeding other compared methods.

\subsection{Overall architecture}
We tackle the classic trade off between receptive field size and efficiency with a layered design that follows prior work \cite{guo2024mambairv2, park2025efficient, lee2025emulating, zhang2024transcending}. First, we build an efficient attention backbone that yields large receptive fields at low cost. On top of this backbone we stack hybrid blocks that integrate local detail and global context, and finally we distill spatial knowledge from large kernel blocks into the network so the model better preserves fine structures and intricate details.

A low resolution input image \(\boldsymbol{I}_{\mathrm{LR}}\in\mathbb{R}^{H\times W\times 3}\) is first mapped to shallow feature embeddings \(\boldsymbol{F}_0\) by a \(3\times 3\) convolution. These shallow features are processed by a core encoder composed of Broad Effective Receptive Field Group (BERFG) units, represented by \(f_{\mathrm{BERFG}}\), to produce deep features \(\boldsymbol{F}_{\mathrm{df}}\). To retain low frequency content, we fuse \(\boldsymbol{F}_{\mathrm{df}}\) with \(\boldsymbol{F}_0\) to obtain \(\boldsymbol{F}_{\mathrm{fuse}}\). The fused features then pass through a reconstruction module (a \(3\times 3\) convolution followed by pixel shuffle \cite{shi2016real}) to yield the final high resolution image \(\boldsymbol{I}_{\mathrm{HR}}\). The core encoder relies on the Broad Effective Receptive Field Group. This group sequentially integrates High Performance Attention for local context, Shared and Receive Hybrid Attention for efficient global aggregation, and Large Kernel Distillation for spatial refinement. We detail these submodules in the following subsections.

\subsection{Broad Effective Receptive Field Group}
To optimize the computational cost of deep attention mechanisms, we propose the Broad Effective Receptive Field Group (BERFG), a two part architecture consisting of a Sharing Block (SB) and a Receiving Block (RB) (Fig.~\ref{figDetailarch}).

The Sharing Block first processes the input features, \(\boldsymbol{F}_{in}\), and passes them through an initial transformation using High Performance Attention (HPA) and Local Module (LM) layers \cite{park2025efficient}. The result then enters a series of Shared Hybrid Attention (\(f_{SHA}\)) modules that perform the full attention computation. As shown in the equation \ref{eqsha}, these modules critically share \(\boldsymbol{A}^{(a)}_{qk}\) and \(\boldsymbol{A}^{(a)}_{map}\) as
\begin{equation}
\boldsymbol{F}_{2, a+1}, \boldsymbol{A}^{(a)}_{map}, \boldsymbol{A}^{(a)}_{qk} = f^{(a)}_{SHA}(\boldsymbol{F}_{2, a})
\label{eqsha}
\end{equation}
The features are subsequently refined by Large Kernel Distillation (LKD) layers. Finally, the SB output, \(\boldsymbol{F}_{SB\_out}\), is produced by an Enhanced Spatial Attention (ESA) module \cite{liu2020rfa} applied after a residual connection with the initial input, \(\boldsymbol{F}_{in}\).

The Receiving Block takes \(\boldsymbol{F}_{SB\_out}\) as its input and mirrors its structure. However, it introduces our core optimization within its Received Hybrid Attention (\(f_{RHA}\)) modules. Instead of recomputing the entire attention mechanism, the RB directly reuses the attention components (\(\boldsymbol{A}^{(a)}_{map}, \boldsymbol{A}^{(a)}_{qk}\)) pre calculated by the corresponding \textit{a}th module in the SB as
\begin{equation}
\boldsymbol{F'}_{2, a+1} = f^{(a)}_{RHA}(\boldsymbol{F'}_{2, a}, \boldsymbol{A}^{(a)}_{map}, \boldsymbol{A}^{(a)}_{qk})
\end{equation}
The architecture concludes with the same LKD and ESA layers as the SB to produce the final output, \(\boldsymbol{F}_{out}\). This semi sharing strategy allows the BERFG to maintain the powerful feature representation of deep attention while substantially lowering the computational burden.

\subsection{High performance attention}
Prior work \cite{lee2025emulating} has shown that enlarging the receptive field improves information aggregation. Therefore, we propose an HPA module to optimize the learned information described as
\begin{equation}
\begin{aligned}
\boldsymbol{F}_{mlp} &= f_{\operatorname{ConvMLP}}\left(f_{\operatorname{LN}}\left(\boldsymbol{X}\right)\right) \\ 
\boldsymbol{F}_{1} &= f_{\operatorname{FWA}}\left(f_{\operatorname{LN}}\left(\boldsymbol{F}_{mlp}\right)\right)
\end{aligned}
\end{equation}
with LN being the normalization and Convolutional layer, ConvMLP being the Multi layer Perceptron with the kernel size of seven, and Flash Attention is for the Large Window Attention. We first apply ConvMLP to capture local context without relying on explicit QKV projections. To efficiently aggregate information over a larger spatial area, we incorporate Window Attention with a \(32 \times 32\) window size. However, standard self attention over large windows is computationally expensive due to its quadratic memory and runtime complexity. To address this, we adopt Flash Attention, which enables exact attention computation with significantly lower memory usage and faster execution, making large window self attention feasible. Nevertheless, the effective receptive field remains constrained. To overcome this, we introduce a hybrid module, as detailed below.

\subsection{Shared and Receive Hybrid Attention}
The core concept of our work is to expand the model effective receptive field while maintaining a balance between high performance and a lightweight design. To achieve this, we propose a solution with two main pillars. First, we introduce a semi sharing mechanism that dramatically reduces computational overhead without sacrificing representational power. Second, we design a Hybrid Attention Module that captures local, global, and channel information effectively. Together, these two innovations allow our model to capture more of the information while staying fast and efficient. The detail of the architecture is shown in Fig.~\ref{figDetailarch}.

\paragraph{Semi sharing mechanism.} 
\vspace{-1em}
To achieve a lightweight yet high performance design, we introduce a semi sharing mechanism across Hybrid Attention (HA) Module pairs. As illustrated in Fig.~\ref{figDetailarch}a, SHA consists of Shared Window Multi Head Attention (Shared WMHA) and Shared Dual Fusion Layer (SDFL), while RHA is composed of Received Window Multi Head Attention (Received WMHA) and Dual Fusion Receiver Layer (DFRL). In this design, the attention map from Shared WMHA is shared with Received WMHA, reducing redundant computation. Unlike standard Softmax attention, which derives the map as \(\text{Softmax}(\boldsymbol{QK}^T)\), the Dual Fusion Layer computes it as \(\phi(\boldsymbol{Q})\phi(\boldsymbol{K})^T\). Within this layer, dynamic feature maps with \(\boldsymbol{Q}\) and \(\boldsymbol{K}\) are recomputed independently at each layer. This dynamic recomputation refreshes feature representations while maintaining efficiency. Overall, the semi sharing mechanism balances representation renewal and computational cost, a trade off validated by our ablation studies.

\paragraph{Local spatial module.}
\vspace{-1em}Following previous study \cite{liang2021swinir}, we employ window based multi head self attention (WMSA) to extract fine local textures (Fig.~\ref{figDetailarch}). This design emphasizes the immediate neighborhood of each pixel, allowing high fidelity local restoration.

\paragraph{Global and channel modules.} 
\vspace{-1em}The goal of our dual fusion layer is to efficiently aggregate global spatial and channel information. First, \(\boldsymbol{Q}, \boldsymbol{K}, \boldsymbol{V} \in \mathbb{R}^{HW\times C/2}\)  are computed over the input token \(\boldsymbol{X} \in \mathbb{R}^{HW\times C}\) according to
\begin{equation}
\boldsymbol{Q} = \boldsymbol{W}_Q\boldsymbol{X}; \quad \boldsymbol{K} = \boldsymbol{W}_K\boldsymbol{X}; \quad \boldsymbol{V} = \boldsymbol{W}_V\boldsymbol{X}; 
\end{equation}
where \(\boldsymbol{W}_Q\),\(\boldsymbol{W}_K\), and \(\boldsymbol{W}_V\) are projection matrices. To reduce redundancy and computation, we project from dimension \(C\) to \(C/2\), which halves the feature channels and lightens the model without sacrificing critical information. The tensors are then fed into the spatial, global, and channel branches.

For the spatial branch, we introduce Hedgehog attention (HgA) depicted in Fig.~\ref{figDetailarch}b. We also integrate a lightweight depthwise convolution into the computation process, which can be formulated as
\begin{equation}
\boldsymbol{F}_{sb}=\phi(\boldsymbol{Q})((\phi(\boldsymbol{K)}^T \boldsymbol{V})+\boldsymbol{W}_d \boldsymbol{V},
\end{equation}
where \(\boldsymbol{W}_d\) denotes a depthwise convolution and \(\phi\) is the HFM. We denote the spatial branch output as \(\boldsymbol{F}_{sp}\). The HFM learns multi scale feature interactions and enhances the model ability to capture complex spatial dependencies. In addition, we use Fourier Feature Map \cite{hua2024fourier} to boost spatial attention performance by informing the pixel relative position in image. A depthwise convolution is included to capture local structure and complement global attention.

For the channel branch, the self attention mechanism in the CA mechanism is performed along the channel dimension as depicted in Fig.~\ref{figDetailarch}c. Following prior channel attention designs, we compute channel wise attention as
\begin{equation}
\boldsymbol{F}_{cb}= \text{softmax}(\boldsymbol{Q}^T\boldsymbol{K})V,
\end{equation}
where \(F_{cb}\) is the output of channel branch. 

Finally, \(F_{sb}\) and \(F_{cb}\) are concatenated to get the final output. In terms of computational complexity, both the global spatial branch and the channel branch have linear complexity in spatial resolution. To be more specific, the number of multiply add operations required by both the global spatial branch and the channel branch is
\begin{equation}
\mathcal{O}_{\text{DFL}} = \underbrace{2C^{2}HW}_{\text{\textcolor{blue}{channel branch}}} + \underbrace{\left(6\,HW\,\frac{C^{2}}{D} + 9\,HW\,C\right)}_{\text{\textcolor{red}{spatial branch}}} .
\end{equation}
where \(D\) denotes the number of heads and the complexity of the channel branch is written in \textcolor{blue}{blue} and the spatial branch is written in \textcolor{red}{red}. This shows that our dual fusion layer achieves \textit{linear complexity} in spatial resolution while still maintaining high efficiency.

\subsection{Large Kernel Distillation}
\label{seclkd}
To further refine features with a focus on a wide range of spatial dependencies, we incorporate a Large Kernel Distillation module (Fig.~\ref{figDetailarch}d). We distill computation onto the most informative channels while preserving full content through a lightweight bypass. We split channels into a fine grained subset \(\boldsymbol{F}_{\mathrm{fg}} \in \mathbb{R}^{H \times W \times C_{\mathrm{fg}}}\)  and a coarse subset \(\boldsymbol{F}_{\mathrm{cg}} \in \mathbb{R}^{H \times W \times (C - C_{\mathrm{fg}})}\) with \(C_{\mathrm{fg}} = \max(C/4,\,16)\)
\begin{equation}
\boldsymbol{F}_{\mathrm{fg}},\, \boldsymbol{F}_{\mathrm{cg}} = \mathrm{Split}(\boldsymbol{X}), \quad \boldsymbol{F}_{b+1} = \mathrm{Concat}\!\big(f_{\text{TFE}}(\boldsymbol{F}_{\mathrm{fg}}),\, \boldsymbol{F}_{\mathrm{cg}}\big).
\end{equation}
The Triple Feature Extraction (TFE) module consists of three parallel branches. A \textit{channel attention branch} extracts channel information, a \textit{local branch} with a \(1 \times 1 \rightarrow 3 \times 3 \rightarrow 1 \times 1\) bottleneck captures fine grained details, and a \textit{hierarchical large kernel branch} uses depthwise and dilated separable convolutions to model long range context. This design achieves high efficiency by restricting heavy computation to \(C_{fg}\) channels, which proportionally reduces computation. Simultaneously, the large kernel path efficiently expands the receptive field using dilation and depthwise factorization. Full architectural details, complexity, and effective receptive field analysis are in the supplementary material.
\section{Experiments}
To validate our proposed UCAN models, this section presents its performance on SR task. The evaluation was conducted across five different test sets. We will describe the methodologies and datasets used in our experiments and present a series of ablation studies to confirm the significance of each proposed module.

\begin{table*}[!ht]
\centering
\caption{Quantitative comparison on \textit{\textbf{lightweight image super-resolution}} with state-of-the-art methods. The best and second-best results are shown in \textbf{bold} and \underline{underlined}, respectively.}
\label{tab:lightSR}
\vspace{-3mm}
\setlength{\tabcolsep}{8pt}
\scalebox{0.70}{
\begin{tabular}{@{}l|c|c|c|cc|cc|cc|cc|cc@{}}
\toprule
 & & & & \multicolumn{2}{c|}{\textbf{Set5}} &
  \multicolumn{2}{c|}{\textbf{Set14}} &
  \multicolumn{2}{c|}{\textbf{BSDS100}} &
  \multicolumn{2}{c|}{\textbf{Urban100}} &
  \multicolumn{2}{c}{\textbf{Manga109}} \\
\multirow{-2}{*}{Method} & \multirow{-2}{*}{scale}& \multirow{-2}{*}{\#param}& \multirow{-2}{*}{MACs} & PSNR  & SSIM   & PSNR  & SSIM   & PSNR  & SSIM   & PSNR  & SSIM   & PSNR  & SSIM   \\ \midrule
% SwinIR-light~\cite{liang2021swinir}& $2\times$  & 910K & {244.2G} 
% & 38.14 & {0.9611} & {33.86} & {0.9206} & {32.31} & {0.9012} & {32.76} & {0.9340} & {39.12} & {0.9783} \\
MambaIR-light~\cite{guo2025mambair} & $2\times$  & 905K & 334.2G
& {38.13} & {0.9610} & {33.95} & {0.9208} & {32.31} & {0.9013} & {32.85} & {0.9349} & {39.20} & {0.9782} \\
% ELAN~\cite{zhang2022efficient} & $2\times$  & 621K & 203.1G
% & 38.17  &0.9611  &33.94 &0.9207  &32.30  &0.9012  &32.76  &0.9340 &39.11 &0.9782 \\
OmniSR~\cite{wang2023omni} &$2\times$ & 772K & 194.5G 
&38.22 &0.9613  &33.98 &0.9210  &32.36 &0.9020 &33.05 &0.9363 &39.28 &0.9784 \\
SRFormer-light~\cite{zhou2023srformer} & $2\times$  & 853K & 236.3G
&38.23 &0.9613 &33.94 &0.9209 &32.36 &0.9019 &32.91 &0.9353 &39.28 &0.9785 \\
ATD-light~\cite{zhang2024transcending} &$2\times$ & 753K    & 380.0G 
& 38.29 & 0.9616  & 34.10 &0.9217  & 32.39 &0.9023  & 33.27 &0.9375  & 39.52 &0.9789  \\
HiT-SRF~\cite{zhang2025hit} &$2\times$ & 847K & 226.5G  
& 38.26 &0.9615  & 34.01 &0.9214  & 32.37 &0.9023  & 33.13 &0.9372  & 39.47 &0.9787 \\
ASID-D8~\cite{park2025efficient} & $2\times$ & 732K & 190.5G$^{\dagger}$  
& 38.32 & \underline{0.9618}  & \underline{34.24} & \underline{0.9232}  & 32.40 & \underline{0.9028}  & 33.35 & \underline{0.9387}  & -~ &~- \\
MambaIRV2-light~\cite{guo2024mambairv2attentivestatespace} &$2\times$ & 774K & 286.3G  
& 38.26 &0.9615  & 34.09 &0.9221  & 32.36 &0.9019  & 33.26 &0.9378 & 39.35 &0.9785 \\
RDN~\cite{zhang2018residual} &$2\times$ & 22123K & 5096.2G 
& 38.24 &0.9614 & 34.01 &0.9212 & 32.34 &0.9017  & 32.89 &0.9353 & 39.18 &0.9780 \\
RCAN~\cite{zhang2018image} &$2\times$ & 15445K & 3529.7G  
& 38.27
&0.9614 
& 34.12
&0.9216 
& \underline{32.41}
& 0.9027
& 33.34
&0.9384 
& 39.44
&0.9786 \\
ESC ~\cite{park2025efficient} &$2\times$ &  947K & 592.0G & \underline{38.35} & \textbf{0.9619}& 34.11 & 0.9223 & \underline{32.41} & 0.9027 & \textbf{33.46} & \textbf{0.9395} & \underline{39.54}& \textbf{0.9790}\\
\textbf{UCAN (Our)} & $2\times$  &  689K & 146.3G 
& 38.34
& \underline{0.9618}
& \textbf{34.27}
& \textbf{0.9242}
& 32.39
& 0.9025	
& 33.22	
& 0.9379	
& \underline{39.54}
& \textbf{0.9790}
\\
\textbf{UCAN-L (Our)}& $2\times$  &  886K & 182.4G 
& \textbf{38.37}
& \textbf{0.9619}
& 34.19
& 0.9224
& \textbf{32.44}
& \textbf{0.9031}
& \underline{33.39}
& \underline{0.9393}
& \textbf{39.66}
& \underline{0.9789}
\\
\midrule
% SwinIR-light~\cite{liang2021swinir} & $3\times$ & 918K & 111.2G & 34.62 & 0.9289 & 30.54 & 0.8463 & 29.20 & 0.8082 & 28.66 & 0.8624 & 33.98 & 0.9478 \\
MambaIR-light~\cite{guo2025mambair} & $3\times$ & 913K & 148.5G & 34.63 & 0.9288 & 30.54 & 0.8459 & 29.23 & 0.8084 & 28.70 & 0.8631 & 34.12 & 0.9479 \\
% ELAN~\cite{zhang2022efficient} & $3\times$ & 629K & 90.1G & 34.61 & 0.9288 & 30.55 & 0.8463 & 29.21 & 0.8081 & 28.69 & 0.8624 & 34.00 & 0.9478 \\
OmniSR~\cite{wang2023omni} & $3\times$ & 780K & 88.4G & 34.70 & 0.9294 & 30.57 & 0.8469 & 29.28 & 0.8094 & 28.84 & 0.8656 & 34.22 & 0.9487 \\
SRformer-light~\cite{zhou2023srformer} & $3\times$ & 861K & 105.4G & 34.67 & 0.9296 & 30.57 & 0.8469 & 29.26 & 0.8099 & 28.81 & 0.8655 & 34.19 & 0.9489 \\
ATD-light~\cite{zhang2024transcending} & $3\times$ & 760K & 168.0G & 34.74 & 0.9300 & 30.68 & 0.8485 & \underline{29.32} & 0.8109 & \underline{29.17} & 0.8709 & 34.60 & 0.9506 \\
HiT-SRF~\cite{zhang2025hit} & $3\times$ & 855K & 101.6G & 34.75 & 0.9300 & 30.61 & 0.8475 & 29.29 & 0.8106 & 28.99 & 0.8687 & 34.53 & 0.9502 \\
ASID-D8~\cite{park2025efficient} & $3\times$ & 739K & 86.4G & \textbf{34.84} & 0.9307 & 30.66 & 0.8491 & \underline{29.32} & \underline{0.8119} & 29.08 & 0.8706 & - & - \\
MambaIRV2-light~\cite{guo2024mambairv2attentivestatespace} & $3\times$ & 781K & 126.7G & 34.71 & 0.9298 & 30.68 & 0.8483 & 29.26 & 0.8098 & 29.01 & 0.8689 & 34.41 & 0.9497 \\
RDN~\cite{zhang2018residual} & $3\times$ & 22308K & 2281.2G & 34.71 & 0.9296 & 30.57 & 0.8468 & 29.26 & 0.8093 & 28.80 & 0.8653 & 34.13 & 0.9484 \\
RCAN~\cite{zhang2018image} & $3\times$ & 15629K & 1586.1G & 34.74 & 0.9299 & 30.65 & 0.8482 & \underline{29.32} & 0.8111 & 29.09 & 0.8702 & 34.44 & 0.9499 \\
ESC~\cite{park2025efficient} & $3\times$ & 955K & 267.6G & \textbf{34.84} & \underline{0.9308} & \underline{30.74} & \underline{0.8493} & \textbf{29.34} & 0.8118 & \underline{29.28} & \textbf{0.8739} & \underline{34.66} & \underline{0.9512} \\
\textbf{UCAN (Our)} & $3\times$ & 696K & 64.6G & \underline{34.83} & \underline{0.9308} & 30.72 & \underline{0.8493} & \underline{29.32} & \underline{0.8121} & 29.15 & 0.8712 & 34.62 & 0.9508 \\
\textbf{UCAN-L(Our)} & $3\times$ & 893K & 81.3G & 34.81 & \textbf{0.9311} & \textbf{30.75} & \textbf{0.8500} & \textbf{29.34} & \textbf{0.8127} & \textbf{29.29} & \underline{0.8738} & \textbf{34.79} & \textbf{0.9516} \\
\midrule
% SwinIR-light~\cite{liang2021swinir} & $4\times$  & 930K & {63.6G} 
% & {32.44}
% & {0.8976}
% & {28.77}
% & {0.7858}
% & {27.69}
% & {0.7406}
% & {26.47}
% & {0.7980}
% & {30.92}
% & {0.9151}
% \\
MambaIR-light~\cite{guo2025mambair} & $4\times$  &  924K & 84.6G 
& {32.42}
& {0.8977}
& {28.74}
& {0.7847}
& {27.68}
& {0.7400}
& {26.52}
& {0.7983}
& {30.94}
& {0.9135}
\\
% ELAN~\cite{zhang2022efficient} & $4\times$  & 640K& 54.1G 
% &32.43 
% &0.8975 
% &28.78 
% &0.7858 
% &27.69 
% &0.7406 
% &26.54 
% &0.7982 
% &30.92
% &0.9150
% \\
OmniSR~\cite{wang2023omni} & $4\times$ & 792K & 50.9G & 32.49 &0.8988 & 28.78 &0.7859 & 27.71 &0.7415 & 26.64 &0.8018 & 31.02 &0.9151 \\
SRformer-light~\cite{zhou2023srformer} & $4\times$ & 873K &62.8G 
& 32.51 
& 0.8988 
& 28.82 
& 0.7872 
& 27.73
& 0.7422 
& 26.67
& 0.8032
& 31.17 
& 0.9165
\\
ATD-light~\cite{zhang2024transcending} &$4\times$ & 769K & 100.1G  & 32.63 &0.8998 & 28.89 &0.7886 & \underline{27.79} &0.7440 & \underline{26.97} &0.8107 & 31.48 &0.9198 \\
HiT-SRF~\cite{zhang2025hit} &$4\times$ & 866K  & 58.0G  & 32.55 &0.8999 & 28.87 &0.7880 & 27.75 &0.7432 & 26.80 &0.8069 & 31.26 &0.9171 \\
ASID-D8~\cite{park2025efficient} &$4\times$ & 748K  & 49.6G  & 32.57 &0.8990 & 28.89 &0.7898 & 27.78 &0.7449 & 26.89 &0.8096 & -~ & ~- \\
MambaIRV2-light~\cite{guo2024mambairv2attentivestatespace} &$4\times$ & 790K  & 75.6G  & 32.51 &0.8992 & 28.84 &0.7878 & 27.75 &0.7426 & 26.82 &0.8079 & 31.24 &0.9182 \\
RDN~\cite{zhang2018residual} &$4\times$  & 22271K & 1309.2G  & 32.47 &0.8990 & 28.81 &0.7871 & 27.72 &0.7419 & 26.61 &0.8028 & 31.00 &0.9151 \\
RCAN~\cite{zhang2018image} &$4\times$  & 15592K & 917.6G  & \underline{32.63} &0.9002 & 28.87 &0.7889 & 27.77 &0.7436 & 26.82 &0.8087 & 31.22 &0.9173 \\
ESC ~\cite{park2025efficient}  &$4\times$ & 968K & 149.2G & \textbf{32.68}& \underline{0.9011} & \underline{28.93} & \underline{0.7902} & \textbf{27.80} & \underline{0.7447} & \textbf{27.07} & \textbf{0.8144} & \underline{31.54} & \underline{0.9207} \\
\textbf{UCAN (Our)}& $4\times$  &  705K & 38.1G 
& \underline{32.65}
& 0.9010
& \underline{28.95}	
& 0.7899
& \underline{27.79}
& \underline{0.7454}	
& 26.89
& 0.8097
& \underline{31.50}	
& 0.9200
\\
\textbf{UCAN-L (Our)}& $4\times$  &  902K & 48.4G 
& \textbf{32.68}
& \textbf{0.9015}
& \textbf{28.99}	
& \textbf{0.7917}
& \textbf{27.80}
& \textbf{0.7459}	
& \underline{27.06}	
& \underline{0.8134}	
& \textbf{31.63}	
& \textbf{0.9212}
\\
\bottomrule
\end{tabular}%
}
\label{tab:comparative_result}
\end{table*}

\subsection{Classic image super-resolution}
For evaluating our proposed architectures on traditional super-resolution tasks, we utilized five standard benchmark datasets (Set5 ~\cite{bevilacqua2012low}, Set14 ~\cite{zeyde2010single}, BSDS100 ~\cite{martin2001database}, Urban100 ~\cite{huang2015single}, Manga109 \cite{matsui2017sketch}) and assessed performance using PSNR and SSIM metrics calculated on the luminance channel following border cropping based on scaling factors and YCbCr color space conversion. Performance analysis involved reconstructing HD images (1280 $\times$ 720), while computational efficiency was measured using multiply-accumulate operations (MACs) via the fvcore library. We present two model variants: UCAN, designed for parameter efficiency, and UCAN-L, optimized for maximizing PSNR and SSIM performance.

\subsection{Training Configuration}
Our model is trained using a batch size of 16, with input images randomly cropped to 64 $\times$ 64 pixels and augmented through random rotation and horizontal flipping operations. The optimization employs the Adam optimizer ($\beta_1 = 0.9$, $\beta_2 = 0.99$) to minimize L1 reconstruction loss, LDL loss \cite{liang2022details} and Wavelet loss \cite{korkmaz2024training}, following established super-resolution practices. For $\times$2 scale factor, we train from scratch for 800,000 iterations with an initial learning rate of $5 \times 10^{-4}$, applying step-wise decay by half at milestones [300K, 500K, 650K, 700K, 750K]. Higher magnification factors ($\times$3 and $\times$4) utilize transfer learning by fine-tuning the pre-trained $\times$2 model for 400,000 iterations, maintaining the same initial learning rate with decay at [200K, 320K, 360K, 380K] iterations. All experiments are implemented in PyTorch on 2 RTX 3090 GPUs.

\begin{figure}[ht!]
    \begin{minipage}{0.85\linewidth}
        \begin{tabular}{@{}c@{\hspace{1mm}}c@{}}
            % --------- Block 1: Urban100 - img73 ---------
            \begin{minipage}[c][0.3\linewidth][c]{0.5\linewidth}
                \centering
                \includegraphics[width=\linewidth]{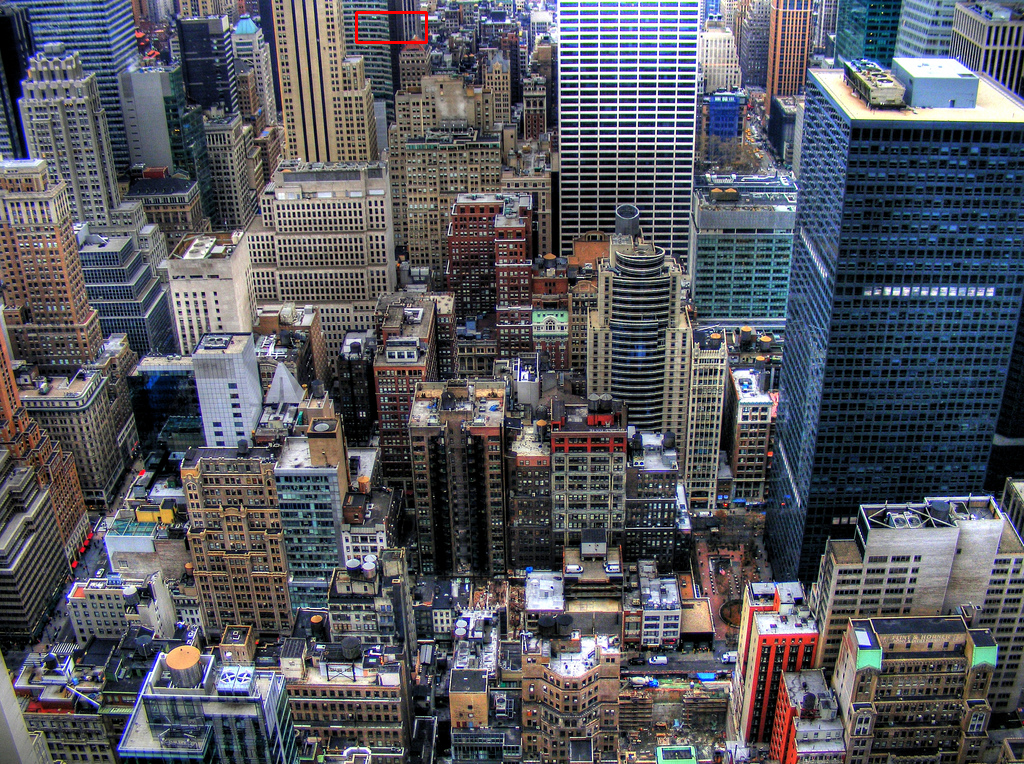} \\
                \vspace{0.5mm}
                {\fontsize{6.5}{7}\selectfont Urban100 - img73}
            \end{minipage}
            &
            \begin{minipage}[c]{0.65\linewidth}
                \centering
                \begin{minipage}[t]{0.32\linewidth}
                    \centering
                    \includegraphics[width=\linewidth]{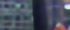} \\
                    {\fontsize{6.25}{7}\selectfont SRFormer \\ 17.78}
                \end{minipage}\hfill
                \begin{minipage}[t]{0.32\linewidth}
                    \centering
                    \includegraphics[width=\linewidth]{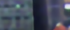} \\
                    {\fontsize{6.25}{7}\selectfont ATD \\ 17.40}
                \end{minipage}\hfill
                \begin{minipage}[t]{0.32\linewidth}
                    \centering
                    \includegraphics[width=\linewidth]{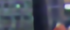} \\
                    {\fontsize{6.25}{7}\selectfont MambaIRV2-lt \\ 16.99}
                \end{minipage}

                \vspace{1mm}

                \begin{minipage}[t]{0.32\linewidth}
                    \centering
                    \includegraphics[width=\linewidth]{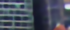} \\
                    {\fontsize{6.25}{7}\selectfont HiT-SRF \\ 19.29}
                \end{minipage}\hfill
                \begin{minipage}[t]{0.32\linewidth}
                    \centering
                    \includegraphics[width=\linewidth]{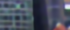} \\
                    {\fontsize{6.25}{7}\selectfont \textbf{UCAN-L} \\ \textbf{18.39}}
                \end{minipage}\hfill
                \begin{minipage}[t]{0.32\linewidth}
                    \centering
                    \includegraphics[width=\linewidth]{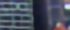} \\
                    {\fontsize{6.25}{7}\selectfont \textbf{UCAN} \\ \textcolor{red}{\textbf{19.70}}}
                \end{minipage}
            \end{minipage}
        \end{tabular}
        \begin{tabular}{@{}c@{\hspace{1mm}}c@{}}
            % --------- Block 2: Urban100 - img44 ---------
            \begin{minipage}[c][0.5\linewidth][c]{0.5\linewidth}
                \centering
                \includegraphics[width=\linewidth]{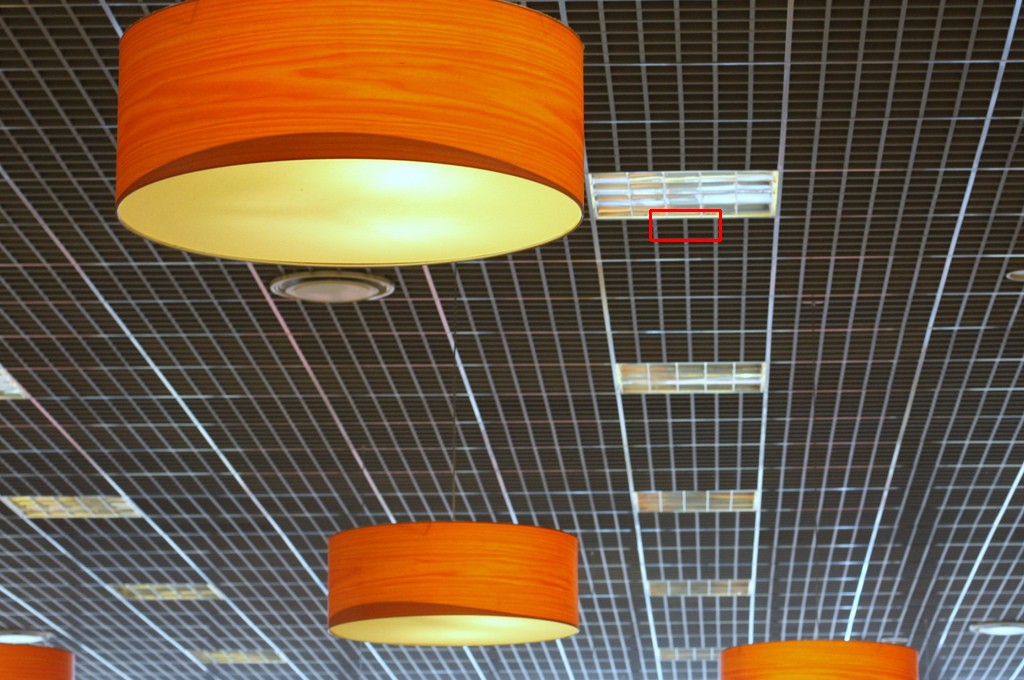} \\
                \vspace{1mm}
                {\fontsize{6.5}{7}\selectfont Urban100 - img44}
            \end{minipage}
            &
            \begin{minipage}[c]{0.65\linewidth}
                \centering
                \begin{minipage}[t]{0.32\linewidth}
                    \centering
                    \includegraphics[width=\linewidth]{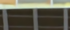} \\
                    {\fontsize{6.25}{7}\selectfont SRFormer \\ 23.76}
                \end{minipage}\hfill
                \begin{minipage}[t]{0.32\linewidth}
                    \centering
                    \includegraphics[width=\linewidth]{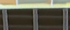} \\
                    {\fontsize{6.25}{7}\selectfont ATD \\ 23.13}
                \end{minipage}\hfill
                \begin{minipage}[t]{0.32\linewidth}
                    \centering
                    \includegraphics[width=\linewidth]{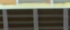} \\
                    {\fontsize{6.25}{7}\selectfont MambaIRV2-lt \\ 25.72}
                \end{minipage}

                \vspace{1mm}

                \begin{minipage}[t]{0.32\linewidth}
                    \centering
                    \includegraphics[width=\linewidth]{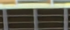} \\
                    {\fontsize{6.25}{7}\selectfont HiT-SRF \\ 23.01}
                \end{minipage}\hfill
                \begin{minipage}[t]{0.32\linewidth}
                    \centering
                    \includegraphics[width=\linewidth]{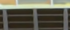} \\
                    {\fontsize{6.25}{7}\selectfont \textbf{UCAN-L} \\ {\textbf{27.36}}}
                \end{minipage}\hfill
                \begin{minipage}[t]{0.32\linewidth}
                    \centering
                    \includegraphics[width=\linewidth]{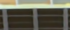} \\
                    {\fontsize{6.25}{7}\selectfont \textbf{UCAN} \\ \textbf{26.35}}
                \end{minipage}
            \end{minipage}
        \end{tabular}
    \end{minipage}
    \vspace{-3mm}
    \caption{Visual comparison between ground truth and different methods on Urban100.}
    \vspace{-3mm}
    \label{fig:visual_comparison_urban100}
\end{figure}

\subsection{Benchmarking with State-of-the-Art Approaches}
In this section, we evaluate the effectiveness of our UCAN, UCAN-L model by comparing it with state-of-the-art (SOTA) methods, including CNN-based, lightweight Transformer-based, and SSM-based super-resolution (SR) approaches. We present both quantitative and qualitative results, highlighting comparisons with previous SR models such as SwinIR-light \cite{liang2021swinir}, ELAN \cite{zhang2022efficient}, OmniSR \cite{wang2023omni}, SRformer-light \cite{zhou2023srformer}, ATD-light \cite{zhang2024transcending}, HiT-SRF \cite{zhang2025hit}, ASID-D8 \cite{park2025efficient}, MambaIR-light\cite{guo2025mambair}, MambaIRV2-light \cite{guo2024mambairv2}, RDN \cite{zhang2018residual}, RCAN\cite{zhang2018image} and ESC.

We present a comprehensive comparison of our models against prior state-of-the-art lightweight SR methods, with visual results in Fig. \ref{fig:visual_comparison_urban100} and quantitative metrics in Table \ref{tab:comparative_result}.

Our base model, UCAN, demonstrates exceptional efficiency. For instance, on the Manga109 (×4) dataset, it achieves a significant 0.26 dB PSNR improvement over MambaIRV2 while requiring 11\% fewer parameters. Similarly, when compared to ASID-D8, UCAN delivers superior performance across the evaluation sets despite having 6\% fewer parameters and 23\% lower MACs.

The larger variant, UCAN-L, further solidifies our lead. Despite having 7\% fewer parameters than the strong ESC model, UCAN-L outperforms it on four out of the five benchmark datasets. This is particularly evident on Manga109 ($\times$2), where our model achieves a notable gain of nearly 0.1 dB. These results collectively underscore UCAN's design as a robust, powerful, and resource-efficient solution for modern image super-resolution.

%Despite having fewer than 0.9 million parameters, UCAN-L consistently delivers competitive performance, often matching or surpassing methods such as ESC and ELAN in both PSNR and SSIM metrics. In particular, UCAN-L achieves the highest SSIM in Manga109 (0.9789) and Urban100 (0.9496), while maintaining strong PSNR values. These results demonstrate the effectiveness of UCAN-L in both standard and texture-rich image datasets. Compared to ESC, which is significantly larger, UCAN-L gives similar or better results but uses much less processing power, highlighting its ability to balance reconstruction quality and efficiency. This  result underscores UCAN as a robust and resource-efficient solution for image super-resolution tasks.
% \subsection{Flash Softmax Attention, Softmax attention and Linear Attention}

\subsection{Efficient Receptive Field Comparison}

\begin{figure}[!b]

\centering
\includegraphics[width=0.85\columnwidth]{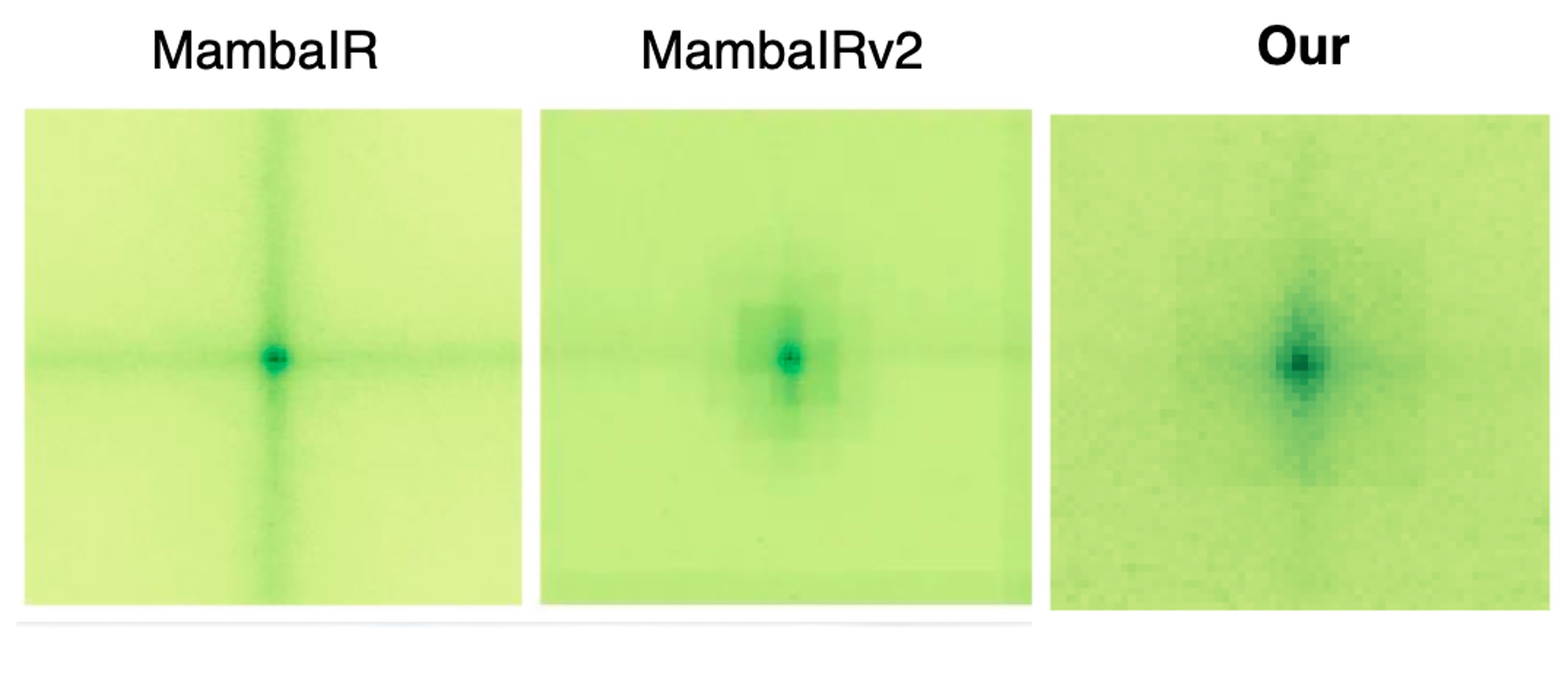} 
% Reduce the figure size so that it is slightly narrower than the column. Don't use precise values for figure width.This setup will avoid overfull boxes.

\caption{Visualize in detail ERF of MambaIR \cite{guo2025mambair}, MambaIRv2\cite{guo2024mambairv2} and UCAN}
\label{fig:ERF}
\vspace{-3mm}
\end{figure}

We present a comprehensive ERF analysis in Fig.~\ref{fig:ERF}, comparing our proposed UCAN model with other models that exploit global spatial information. The ERF images show that UCAN exhibits significantly darker and more extended regions of influence than existing methods. This improved receptive field coverage indicates that our model effectively captures broader contextual information from the input, leading to superior feature representation and ultimately achieving better image reconstruction performance.

\subsection{Local Attribution Maps}
\begin{figure}[h!]
    \centering
    \subcaptionbox*{Input image}[0.22\textwidth]{\includegraphics[width=\linewidth]{sec/IMG/window_position.png}}
    \hfill
    \subcaptionbox*{SRFormer}[0.22\textwidth]{\includegraphics[width=\linewidth]{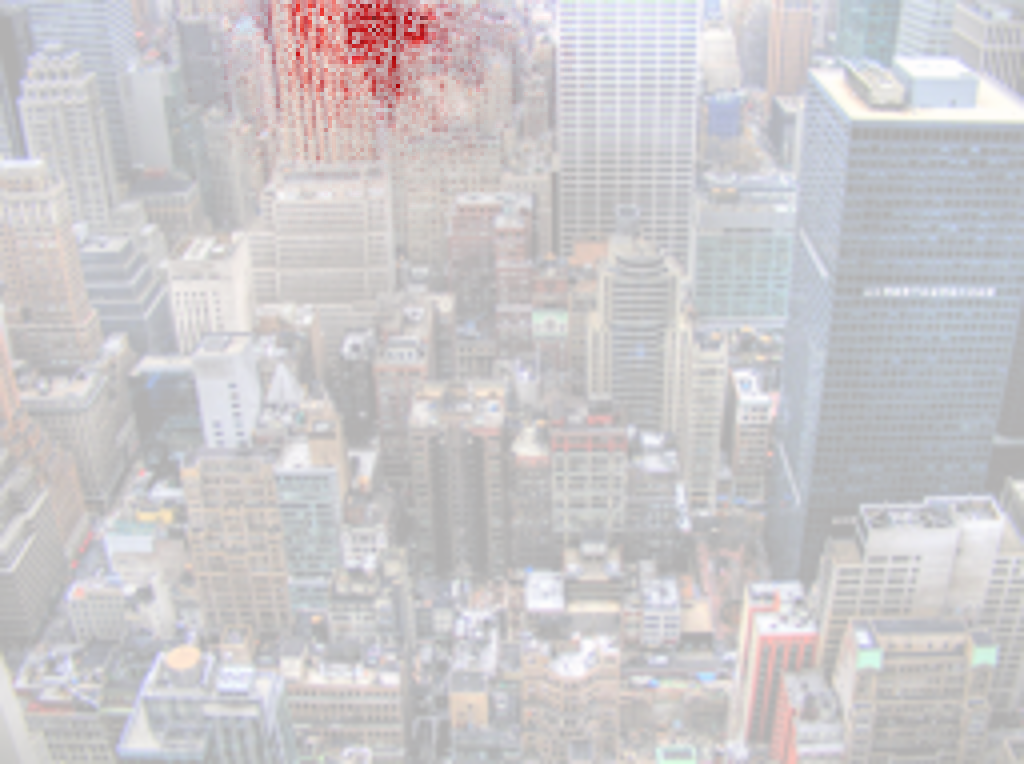}}
    
    \vspace{0.8em} % space between rows

    \subcaptionbox*{ATD}[0.22\textwidth]{\includegraphics[width=\linewidth]{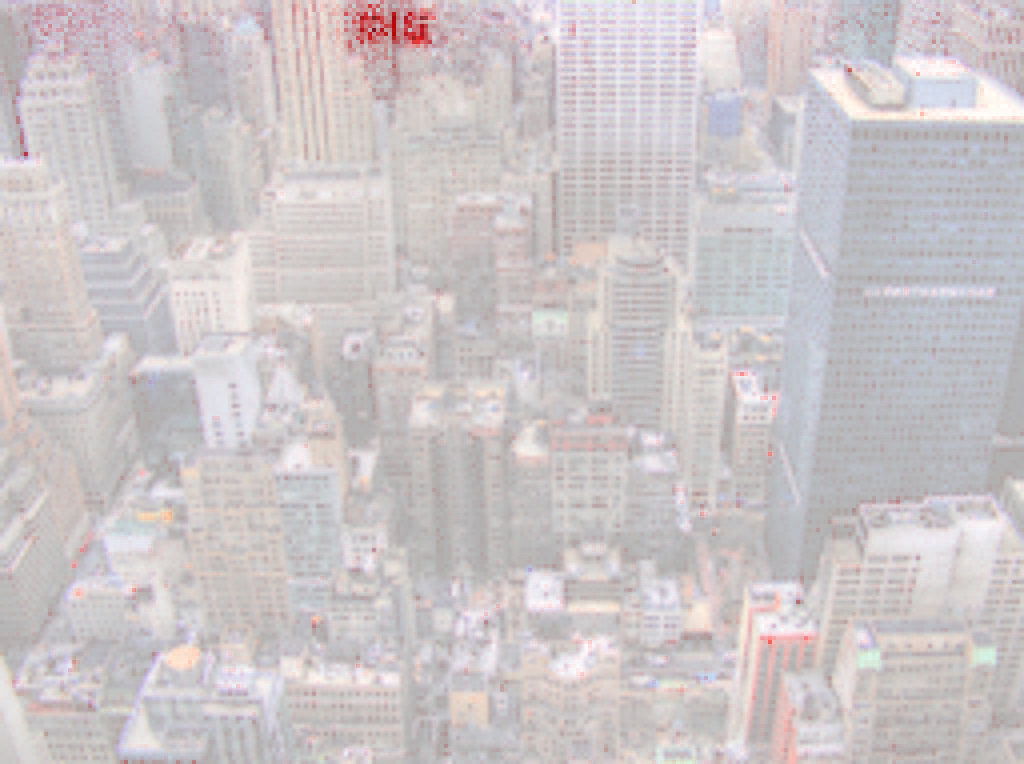}}
    \hfill
    \subcaptionbox*{UCAN (ours)}[0.22\textwidth]{\includegraphics[width=\linewidth]{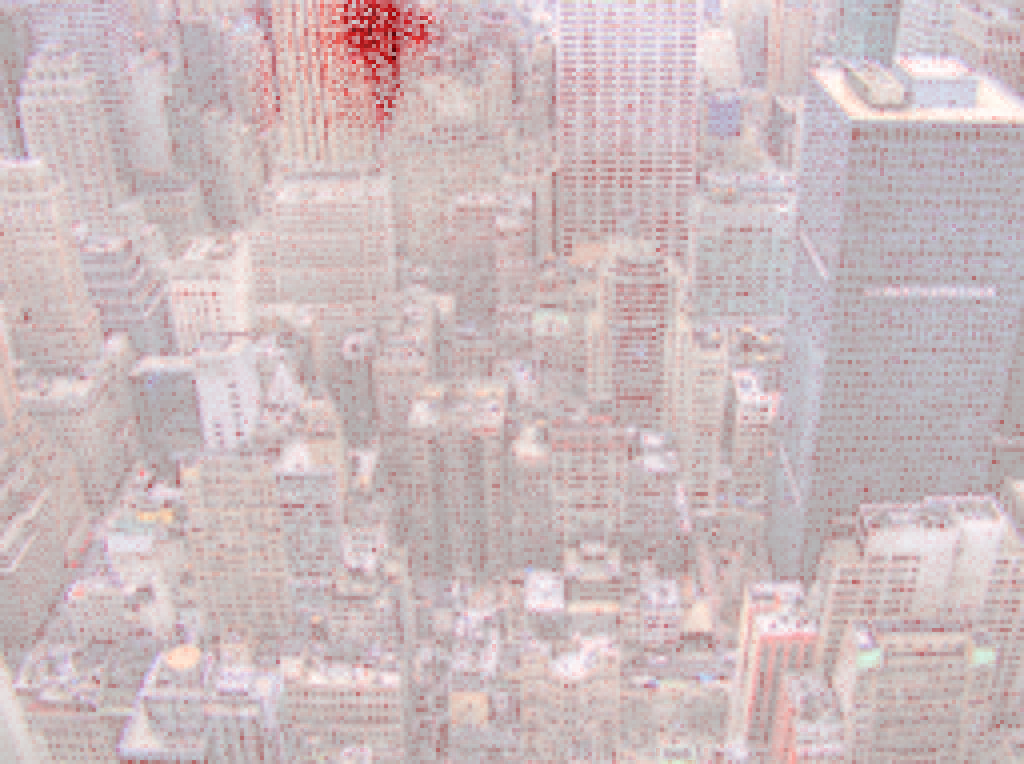}}
    \vspace{-3mm}
    \caption{Local attribution maps (LAM) comparison of different super-resolution methods at scale $\times$4.}
    \vspace{-3mm}
    \label{fig:LamAttribution}
\end{figure}

To assess the information aggregation performance of UCAN, we utilize Local Attribution Maps (LAM) \cite{gu2021interpreting}. LAM is a technique designed to identify which pixels in an input image most strongly influence the final SR output. These regions are known as informative areas, and their size corresponds to the ability of the model to aggregate information. As illustrated in Fig. \ref{fig:LamAttribution}, we applied the LAM method to SRFormer \cite{zhou2023srformer}, ATD \cite{zhang2024transcending}, HiT-SRF \cite{zhang2025hit} and UCAN to compare their respective informative areas for identical target regions. SRFormer operates with small, fixed windows that produce locally confined informative areas, thereby limiting its ability to capture long-range dependencies. While ATD and UCAN have expanded these information domains through global attention mechanisms, our model demonstrates superior information aggregation capabilities. Consequently, our method can leverage information from significantly broader contexts, including repetitive patterns and similar structures across the image, leading to enhanced super-resolution performance.

\subsection{Performance analysis on Attention schemes}
\vspace{-1.5mm}
\begin{table}[ht]
    \centering
    \caption{Comparison of Latency and Parameters between Naive Self-Attention, Flash Attention \cite{dao2022flashattention} , and the Dual Fusion Layer (DFL)}
    \scalebox{0.77}{%
    \begin{tabular}{lcccc}
        \toprule
        \multirow{2}{*}{Module} & \multicolumn{2}{c}{$\mathbb{R}^{64\times64\times64}$} & \multicolumn{2}{c}{$\mathbb{R}^{128\times128\times64}$} \\
        \cmidrule(lr){2-3} \cmidrule(lr){4-5}
        & Latency (ms) & \#params & Latency (ms) & \#params \\
        \midrule
        Attention & \( 596.53 \left( \text{\scriptsize 1.5×} \right) \)& 0.033M & \( 2576.75  \left( \text{\scriptsize 1.0×} \right)\) & 0.082M \\
        Flash Attention & \( 190.50 \left( \text{\scriptsize 4.7×} \right) \) & & \( 191.80  \left( \text{\scriptsize \textcolor{red}{13.4×}} \right) \) & \\
        \cmidrule(lr){1-5}
        DFL & \( 903.47 \left( \text{\scriptsize 1.0×} \right) \) & 0.014M & \( 1294.83  \left( \text{\scriptsize 2.0×} \right) \) & 0.014M \\
        \bottomrule
    \end{tabular}
    }
    \label{tab:attentionModuleComparation}
\end{table}
Efficient long-range interaction is vital for high-resolution super-resolution, yet conventional self-attention struggles with rapidly increasing memory and latency as resolution grows. UCAN overcomes this limitation by combining Flash Attention, which handles large-window modeling efficiently, with a DLF that aggregates global context at constant complexity. Table \ref{tab:attentionModuleComparation} presents a comparison of latency and parameter counts for Native Self-Attention, Flash Attention, and DFL measured over 1000 images on an A100 GPU. Flash Attention achieves up to $13.4\times$ faster processing than vanilla attention at $128 \times 128$ spatial resolution, demonstrating excellent scalability when memory constraints tighten. At smaller resolutions such as $64 \times 64$, the DFL shows slightly higher latency, yet it becomes about twice as fast at larger scales, confirming its efficiency for high-resolution reconstruction. Its parameter count remains stable across resolutions, unlike self-attention whose cost grows with feature dimension, showing that the hybrid design of UCAN balances local efficiency and global awareness effectively.

\subsection{Ablation study}

First, we examined the HPA block under two configurations: complete removal and a regular $16\times16$ window. Both variants led to noticeable performance degradation compared to UCAN, confirming that our default $32\times32$ setting provides the optimal balance.
Next, we compared an $8\times8$ window with the standard $16\times16$ configuration in the WMHA block. The smaller window resulted in a substantial performance drop, indicating that further reduction in window size compromises feature aggregation and overall accuracy.
We then varied the kernel size (KS) to $5\times5$ and $47\times47$ to determine its effect. The $5\times5$ setting produced a smaller receptive field, limiting contextual awareness, while the $47\times47$ setting introduced excessive padding that degraded performance. In both cases, due to the distillation-based architecture, we observed a marked drop in accuracy on Urban100, which contains highly detailed textures.
Finally, we evaluated the semi-sharing mechanism against full sharing (i.e., parameters computed once and reused across layers). The semi-sharing approach achieved better results, confirming that refreshing new information across layers is essential for maintaining representation diversity.

% \usepackage{multirow}
% \usepackage{graphicx}
\begin{table}[ht!]
\centering
\vspace{-3mm}
\caption{Ablation study. We train all models on DIV2K for 400K iterations, and test on Set5 and Urban100 ($\times$2). The final result is shown in the last row.}
\vspace{-3mm}
\label{tab_ablation_table}
\scalebox{0.76}{%
\begin{tabular}{cccccc}
\hline
\multirow{2}{*}{Block}                                                                       & \multirow{2}{*}{Case} & \multicolumn{2}{c}{Set5}                               & \multicolumn{2}{c}{Urban100}                           \\ \cline{3-6} 
                                                                                             &                       & PSNR                      & SSIM                       & PSNR                      & SSIM                       \\ \hline
\multirow{2}{*}{\begin{tabular}[c]{@{}c@{}}High performance \\ attention (HPA)\end{tabular}} & w/o HPA               & 38.27                     & 0.9616                     & 32.90                     & 0.9346                     \\
                                                                                             & window $16\times16$   & 38.32                     & 0.9617                     & 33.04                     & 0.9364                     \\ \hline
WMHA                                                                                         & WS $8\times8$         & 38.32                     & 0.9617                     & 33.02                     & 0.9361                     \\ \hline
\multirow{2}{*}{\begin{tabular}[c]{@{}c@{}}Dual Fusion \\ Layer (DFL)\end{tabular}}          & Using ReLU            & 38.33                     & 0.9618                     & 33.16                     & 0.9374                     \\
                                                                                             & Using ELU + 1         & 38.33                     & 0.9618                     & 33.16                     & 0.9373                     \\ \hline
\multirow{2}{*}{\begin{tabular}[c]{@{}c@{}}Large Kernel \\ Distillation (LKD)\end{tabular}}  & KS = $5\times5$       & \multicolumn{1}{l}{38.33} & \multicolumn{1}{l}{0.9618} & \multicolumn{1}{l}{33.12} & \multicolumn{1}{l}{0.9369} \\
                                                                                             & KS = $47\times47$     & 38.34                     & 0.9618                     & 33.15                     & 0.9372                     \\ \hline
Sharing                                                                                      & Sharing Full          & 38.29                     & 0.9617                     & 32.89                     & 0.9350                     \\ \hline
UCAN                                                                                         & Our                   & 38.34                     & 0.9618                     & 33.22                     & 0.9379                     \\ \hline
\end{tabular}%
}
\end{table}
\vspace{-2mm}

\section{Conclusion}
\vspace{-2mm}
In this paper, we propose UCAN, a novel and lightweight network for image super-resolution that targets the significant memory and computational demands of Transformer-based models. UCAN strategically expands the receptive field with enhanced Flash and Linear Attention, which our experiments show is critical for performance. Furthermore, we introduce an efficient parameter-sharing scheme together with a distillation-based large-kernel convolution module to ensure efficiency. This dual approach reduces computational load while boosting reconstruction quality. As a result, UCAN harnesses Transformer capacity for high-fidelity image reconstruction while remaining lightweight and practical, establishing an effective direction for efficient SR networks

\section*{Acknowledgement}
We acknowledge Ho Chi Minh City University of Technology (HCMUT), VNU-HCM for supporting this study.\\
This work was also partially supported by National Natural Science Foundation of China (62573295).
{
    \small
    \bibliographystyle{ieeenat_fullname}
    \bibliography{main}

@String(ICCV= {Int. Conf. Comput. Vis.})

@String(ECCV= {Eur. Conf. Comput. Vis.})

@String(AAAI = {AAAI})

@String(ICCV  = {ICCV})

@String(ECCV  = {ECCV})

@article{zhang2024hedgehog,
  title={The hedgehog \& the porcupine: Expressive linear attentions with softmax mimicry},
  author={Zhang, Michael and Bhatia, Kush and Kumbong, Hermann and R{\'e}, Christopher},
  journal={arXiv preprint arXiv:2402.04347},
  year={2024}
}

@inproceedings{guo2025mambair,
  title={Mambair: A simple baseline for image restoration with state-space model},
  author={Guo, Hang and Li, Jinmin and Dai, Tao and Ouyang, Zhihao and Ren, Xudong and Xia, Shu-Tao},
  booktitle={European Conference on Computer Vision},
  pages={222--241},
  year={2025},
  organization={Springer}
}

@misc{guo2024mambairv2attentivestatespace,
      title={MambaIRv2: Attentive State Space Restoration}, 
      author={Hang Guo and Yong Guo and Yaohua Zha and Yulun Zhang and Wenbo Li and Tao Dai and Shu-Tao Xia and Yawei Li},
      year={2024},
      eprint={2411.15269},
      archivePrefix={arXiv},
      primaryClass={eess.IV},
      url={https://arxiv.org/abs/2411.15269}, 
}

@inproceedings{dong2014learning,
  title={Learning a deep convolutional network for image super-resolution},
  author={Dong, Chao and Loy, Chen Change and He, Kaiming and Tang, Xiaoou},
  booktitle={Computer Vision--ECCV 2014: 13th European Conference, Zurich, Switzerland, September 6-12, 2014, Proceedings, Part IV 13},
  pages={184--199},
  year={2014},
  organization={Springer}
}

@article{sun2022shufflemixer,
  title={Shufflemixer: An efficient convnet for image super-resolution},
  author={Sun, Long and Pan, Jinshan and Tang, Jinhui},
  journal={Advances in Neural Information Processing Systems},
  volume={35},
  pages={17314--17326},
  year={2022}
}

@article{dao2022flashattention,
  title={Flashattention: Fast and memory-efficient exact attention with io-awareness},
  author={Dao, Tri and Fu, Dan and Ermon, Stefano and Rudra, Atri and R{\'e}, Christopher},
  journal={Advances in neural information processing systems},
  volume={35},
  pages={16344--16359},
  year={2022}
}

@inproceedings{xie2023large,
  title={Large kernel distillation network for efficient single image super-resolution},
  author={Xie, Chengxing and Zhang, Xiaoming and Li, Linze and Meng, Haiteng and Zhang, Tianlin and Li, Tianrui and Zhao, Xiaole},
  booktitle={Proceedings of the IEEE/CVF conference on computer vision and pattern recognition},
  pages={1283--1292},
  year={2023}
}

@article{wu2024transforming,
  title={Transforming image super-resolution: a convformer-based efficient approach},
  author={Wu, Gang and Jiang, Junjun and Jiang, Junpeng and Liu, Xianming},
  journal={IEEE Transactions on Image Processing},
  year={2024},
  publisher={IEEE}
}

@article{dosovitskiy2020image,
  title={An image is worth 16x16 words: Transformers for image recognition at scale},
  author={Dosovitskiy, Alexey},
  journal={arXiv preprint arXiv:2010.11929},
  year={2020}
}

@inproceedings{liu2021swin,
  title={Swin transformer: Hierarchical vision transformer using shifted windows},
  author={Liu, Ze and Lin, Yutong and Cao, Yue and Hu, Han and Wei, Yixuan and Zhang, Zheng and Lin, Stephen and Guo, Baining},
  booktitle={Proceedings of the IEEE/CVF international conference on computer vision},
  pages={10012--10022},
  year={2021}
}

@article{shi2024vmambair,
  title={Vmambair: Visual state space model for image restoration},
  author={Shi, Yuan and Xia, Bin and Jin, Xiaoyu and Wang, Xing and Zhao, Tianyu and Xia, Xin and Xiao, Xuefeng and Yang, Wenming},
  journal={arXiv preprint arXiv:2403.11423},
  year={2024}
}

@article{guo2024mambairv2,
  title={MambaIRv2: Attentive State Space Restoration},
  author={Guo, Hang and Guo, Yong and Zha, Yaohua and Zhang, Yulun and Li, Wenbo and Dai, Tao and Xia, Shu-Tao and Li, Yawei},
  journal={arXiv preprint arXiv:2411.15269},
  year={2024}
}

@inproceedings{zhang2025hit,
  title={HiT-SR: Hierarchical transformer for efficient image super-resolution},
  author={Zhang, Xiang and Zhang, Yulun and Yu, Fisher},
  booktitle={European Conference on Computer Vision},
  pages={483--500},
  year={2024},
  organization={Springer}
}

@inproceedings{zamir2022restormer,
  title={Restormer: Efficient transformer for high-resolution image restoration},
  author={Zamir, Syed Waqas and Arora, Aditya and Khan, Salman and Hayat, Munawar and Khan, Fahad Shahbaz and Yang, Ming-Hsuan},
  booktitle={Proceedings of the IEEE/CVF conference on computer vision and pattern recognition},
  pages={5728--5739},
  year={2022}
}

@inproceedings{zhou2023srformer,
  title={Srformer: Permuted self-attention for single image super-resolution},
  author={Zhou, Yupeng and Li, Zhen and Guo, Chun-Le and Bai, Song and Cheng, Ming-Ming and Hou, Qibin},
  booktitle={Proceedings of the IEEE/CVF International Conference on Computer Vision},
  pages={12780--12791},
  year={2023}
}

@article{lee2025emulating,
  title={Emulating Self-attention with Convolution for Efficient Image Super-Resolution},
  author={Lee, Dongheon and Yun, Seokju and Ro, Youngmin},
  journal={arXiv preprint arXiv:2503.06671},
  year={2025}
}

@inproceedings{shi2016real,
  title={Real-time single image and video super-resolution using an efficient sub-pixel convolutional neural network},
  author={Shi, Wenzhe and Caballero, Jose and Husz{\'a}r, Ferenc and Totz, Johannes and Aitken, Andrew P and Bishop, Rob and Rueckert, Daniel and Wang, Zehan},
  booktitle={Proceedings of the IEEE conference on computer vision and pattern recognition},
  pages={1874--1883},
  year={2016}
}

@inproceedings{liang2021swinir,
  title={Swinir: Image restoration using swin transformer},
  author={Liang, Jingyun and Cao, Jiezhang and Sun, Guolei and Zhang, Kai and Van Gool, Luc and Timofte, Radu},
  booktitle={Proceedings of the IEEE/CVF international conference on computer vision},
  pages={1833--1844},
  year={2021}
}

@inproceedings{zhang2022efficient,
  title={Efficient long-range attention network for image super-resolution},
  author={Zhang, Xindong and Zeng, Hui and Guo, Shi and Zhang, Lei},
  booktitle={European conference on computer vision},
  pages={649--667},
  year={2022},
  organization={Springer}
}

@inproceedings{wang2023omni,
  title={Omni aggregation networks for lightweight image super-resolution},
  author={Wang, Hang and Chen, Xuanhong and Ni, Bingbing and Liu, Yutian and Liu, Jinfan},
  booktitle={Proceedings of the IEEE/CVF Conference on Computer Vision and Pattern Recognition},
  pages={22378--22387},
  year={2023}
}

@inproceedings{zhang2024transcending,
  title={Transcending the limit of local window: Advanced super-resolution transformer with adaptive token dictionary},
  author={Zhang, Leheng and Li, Yawei and Zhou, Xingyu and Zhao, Xiaorui and Gu, Shuhang},
  booktitle={Proceedings of the IEEE/CVF conference on computer vision and pattern recognition},
  pages={2856--2865},
  year={2024}
}

@inproceedings{park2025efficient,
  title={Efficient attention-sharing information distillation transformer for lightweight single image super-resolution},
  author={Park, Karam and Soh, Jae Woong and Cho, Nam Ik},
  booktitle={Proceedings of the AAAI Conference on Artificial Intelligence},
  volume={39},
  number={6},
  pages={6416--6424},
  year={2025}
}

@inproceedings{zhang2018residual,
  title={Residual dense network for image super-resolution},
  author={Zhang, Yulun and Tian, Yapeng and Kong, Yu and Zhong, Bineng and Fu, Yun},
  booktitle={Proceedings of the IEEE conference on computer vision and pattern recognition},
  pages={2472--2481},
  year={2018}
}

@inproceedings{zhang2018image,
  title={Image super-resolution using very deep residual channel attention networks},
  author={Zhang, Yulun and Li, Kunpeng and Li, Kai and Wang, Lichen and Zhong, Bineng and Fu, Yun},
  booktitle={Proceedings of the European conference on computer vision (ECCV)},
  pages={286--301},
  year={2018}
}

@article{bevilacqua2012low,
  title={Low-complexity single-image super-resolution based on nonnegative neighbor embedding},
  author={Bevilacqua, Marco and Roumy, Aline and Guillemot, Christine and Alberi-Morel, Marie Line},
  year={2012},
  publisher={BMVA press}
}

@inproceedings{zeyde2010single,
  title={On single image scale-up using sparse-representations},
  author={Zeyde, Roman and Elad, Michael and Protter, Matan},
  booktitle={International conference on curves and surfaces},
  pages={711--730},
  year={2010},
  organization={Springer}
}

@inproceedings{martin2001database,
  title={A database of human segmented natural images and its application to evaluating segmentation algorithms and measuring ecological statistics},
  author={Martin, David and Fowlkes, Charless and Tal, Doron and Malik, Jitendra},
  booktitle={Proceedings eighth IEEE international conference on computer vision. ICCV 2001},
  volume={2},
  pages={416--423},
  year={2001},
  organization={IEEE}
}

@inproceedings{huang2015single,
  title={Single image super-resolution from transformed self-exemplars},
  author={Huang, Jia-Bin and Singh, Abhishek and Ahuja, Narendra},
  booktitle={Proceedings of the IEEE conference on computer vision and pattern recognition},
  pages={5197--5206},
  year={2015}
}

@article{matsui2017sketch,
  title={Sketch-based manga retrieval using manga109 dataset},
  author={Matsui, Yusuke and Ito, Kota and Aramaki, Yuji and Fujimoto, Azuma and Ogawa, Toru and Yamasaki, Toshihiko and Aizawa, Kiyoharu},
  journal={Multimedia tools and applications},
  volume={76},
  number={20},
  pages={21811--21838},
  year={2017},
  publisher={Springer}
}

@inproceedings{tai2017image,
  title={Image super-resolution via deep recursive residual network},
  author={Tai, Ying and Yang, Jian and Liu, Xiaoming},
  booktitle={Proceedings of the IEEE conference on computer vision and pattern recognition},
  pages={3147--3155},
  year={2017}
}

@inproceedings{kim2016deeply,
  title={Deeply-recursive convolutional network for image super-resolution},
  author={Kim, Jiwon and Lee, Jung Kwon and Lee, Kyoung Mu},
  booktitle={Proceedings of the IEEE conference on computer vision and pattern recognition},
  pages={1637--1645},
  year={2016}
}

@inproceedings{dong2015compression,
  title={Compression artifacts reduction by a deep convolutional network},
  author={Dong, Chao and Deng, Yubin and Loy, Chen Change and Tang, Xiaoou},
  booktitle={Proceedings of the IEEE international conference on computer vision},
  pages={576--584},
  year={2015}
}

@inproceedings{lim2017enhanced,
  title={Enhanced deep residual networks for single image super-resolution},
  author={Lim, Bee and Son, Sanghyun and Kim, Heewon and Nah, Seungjun and Mu Lee, Kyoung},
  booktitle={Proceedings of the IEEE conference on computer vision and pattern recognition workshops},
  pages={136--144},
  year={2017}
}

@inproceedings{hui2018fast,
  title={Fast and accurate single image super-resolution via information distillation network},
  author={Hui, Zheng and Wang, Xiumei and Gao, Xinbo},
  booktitle={Proceedings of the IEEE conference on computer vision and pattern recognition},
  pages={723--731},
  year={2018}
}

@inproceedings{liang2022details,
  title={Details or artifacts: A locally discriminative learning approach to realistic image super-resolution},
  author={Liang, Jie and Zeng, Hui and Zhang, Lei},
  booktitle={Proceedings of the IEEE/CVF conference on computer vision and pattern recognition},
  pages={5657--5666},
  year={2022}
}

@inproceedings{korkmaz2024training,
  title={Training transformer models by wavelet losses improves quantitative and visual performance in single image super-resolution},
  author={Korkmaz, Cansu and Tekalp, A Murat},
  booktitle={Proceedings of the IEEE/CVF Conference on Computer Vision and Pattern Recognition},
  pages={6661--6670},
  year={2024}
}

@inproceedings{ranftl2021vision,
  title={Vision transformers for dense prediction},
  author={Ranftl, Ren{\'e} and Bochkovskiy, Alexey and Koltun, Vladlen},
  booktitle={Proceedings of the IEEE/CVF international conference on computer vision},
  pages={12179--12188},
  year={2021}
}

@article{lau2024large,
  title={Large separable kernel attention: Rethinking the large kernel attention design in cnn},
  author={Lau, Kin Wai and Po, Lai-Man and Rehman, Yasar Abbas Ur},
  journal={Expert Systems with Applications},
  volume={236},
  pages={121352},
  year={2024},
  publisher={Elsevier}
}

@inproceedings{szegedy2016rethinking,
  title={Rethinking the inception architecture for computer vision},
  author={Szegedy, Christian and Vanhoucke, Vincent and Ioffe, Sergey and Shlens, Jon and Wojna, Zbigniew},
  booktitle={Proceedings of the IEEE conference on computer vision and pattern recognition},
  pages={2818--2826},
  year={2016}
}

@inproceedings{lu2022transformer,
  title={Transformer for single image super-resolution},
  author={Lu, Zhisheng and Li, Juncheng and Liu, Hong and Huang, Chaoyan and Zhang, Linlin and Zeng, Tieyong},
  booktitle={Proceedings of the IEEE/CVF conference on computer vision and pattern recognition},
  pages={457--466},
  year={2022}
}

@inproceedings{gu2021interpreting,
  title={Interpreting super-resolution networks with local attribution maps},
  author={Gu, Jinjin and Dong, Chao},
  booktitle={Proceedings of the IEEE/CVF conference on computer vision and pattern recognition},
  pages={9199--9208},
  year={2021}
}

@inproceedings{liu2020rfa,
  title={Residual feature aggregation network for image super-resolution},
  author={Liu, Jie and Zhang, Wenjie and Tang, Yuting and Tang, Jie and Wu, Gangshan},
  booktitle={Proceedings of the IEEE/CVF conference on computer vision and pattern recognition},
  pages={2359--2368},
  year={2020}
}

@article{hua2024fourier,
  title={Fourier Position Embedding: Enhancing Attention's Periodic Extension for Length Generalization},
  author={Hua, Ermo and Jiang, Che and Lv, Xingtai and Zhang, Kaiyan and Sun, Youbang and Fan, Yuchen and Zhu, Xuekai and Qi, Biqing and Ding, Ning and Zhou, Bowen},
  journal={arXiv preprint arXiv:2412.17739},
  year={2024}
}

@article{ai2025breaking,
  title={Breaking Complexity Barriers: High-Resolution Image Restoration with Rank Enhanced Linear Attention},
  author={Ai, Yuang and Huang, Huaibo and Wu, Tao and Fan, Qihang and He, Ran},
  journal={arXiv preprint arXiv:2505.16157},
  year={2025}
}

@inproceedings{tian2024image,
  title={Image processing gnn: Breaking rigidity in super-resolution},
  author={Tian, Yuchuan and Chen, Hanting and Xu, Chao and Wang, Yunhe},
  booktitle={Proceedings of the IEEE/CVF conference on computer vision and pattern recognition},
  pages={24108--24117},
  year={2024}
}

@inproceedings{long2025progressive,
  title={Progressive Focused Transformer for Single Image Super-Resolution},
  author={Long, Wei and Zhou, Xingyu and Zhang, Leheng and Gu, Shuhang},
  booktitle={Proceedings of the Computer Vision and Pattern Recognition Conference},
  pages={2279--2288},
  year={2025}
}
}

\clearpage
\setcounter{page}{1}
\setcounter{section}{0}
\renewcommand{\thesection}{\Alph{section}}
\maketitlesupplementary

\section{More details about Large Kernel Distillation}

We provide a detailed architectural breakdown of our proposed block, including formal definitions of its parallel branches, a rigorous derivation of the Effective Receptive Field (ERF) for the large-kernel branch, and the final feature fusion strategy.

Let the input feature map be $\boldsymbol{X} \in \mathbb{R}^{H \times W \times C}$. 
The block processes $\boldsymbol{X}$ via three parallel branches.

\subsection{Parallel Branch Definitions}

\textbf{Hierarchical Large Kernel (HLK) Branch.}
This branch captures long-range dependencies. We first define its core building blocks:
\begin{itemize}
    \item A \textbf{separable depthwise convolution block}, $\mathcal{S}(\cdot)$, with kernel size $k$:
    $$
    \mathcal{S}(\boldsymbol{X}, k) = f_{dw}^{k \times 1}\!\left(f_{dw}^{1 \times k}(\boldsymbol{X})\right)
    $$
    \item A \textbf{dilated separable depthwise convolution block}, $\mathcal{S}_{d}(\cdot)$, with kernel $k$ and dilation $d$:
    $$
    \mathcal{S}_{d}(\boldsymbol{X}, k, d) = 
    f_{dw}^{k \times 1,\, d}\!\left(f_{dw}^{1 \times k,\, d}(\boldsymbol{X})\right)
    $$
\end{itemize}
The HLK branch has two configurations, both built as a stack. The first stage is always $\mathcal{S}(\boldsymbol{X}, k_{core})$, which forms a dense feature core.

\begin{enumerate}
    \item \textbf{Standard Configuration ($\mathcal{F}_{LK\text{-S}}$):} This is a two-stage stack, used for smaller receptive fields.
    \begin{equation}
    \mathcal{F}_{LK\text{-S}}(\boldsymbol{X}) = \mathcal{S}_{d}(\mathcal{S}(\boldsymbol{X}, k_{core}), k_{core}, d)
    \end{equation}
    
    \item \textbf{Large Configuration ($\mathcal{F}_{LK\text{-L}}$):} To achieve maximum receptive fields, this configuration extends the standard block into a three-stage stack. The first two stages are identical to the Standard Configuration, after which a third dilated separable depthwise convolution block using $k_{extra}$ is appended.
    \begin{equation}
    \mathcal{F}_{LK\text{-L}}(\boldsymbol{X}) = \mathcal{S}_{d}(\mathcal{S}_{d}(\mathcal{S}(\boldsymbol{X}, k_{core}), k_{core}, d), k_{extra}, d)
    \end{equation}
\end{enumerate} 
The final output is $\boldsymbol{X}_{lk}^{\text{out}} = \mathcal{F}_{LK}(\boldsymbol{X})$, where $\mathcal{F}_{LK}$ is either $\mathcal{F}_{LK\text{-S}}$ or $\mathcal{F}_{LK\text{-L}}$.

\paragraph{Channel Branch (CB).}
The channel branch first computes channel-wise attention using a simple linear projection to weigh the importance of each feature map:
\begin{equation}
    \boldsymbol{X}_c = f_{Linear}(\boldsymbol{X}),
\end{equation}
This $\boldsymbol{X}_c$ is the output of channel branch.

\paragraph{Local Context (LC) Branch.}
This branch, denoted $\mathcal{F}_{LC}(\boldsymbol{X})$, captures fine-grained local details using a bottleneck (hourglass-style) block. Given a channel reduction factor $r$:
\begin{equation}
\begin{aligned}
\boldsymbol{X}_l^{(1)} &= f_{\text{GELU}}\big(f_{\text{Conv}}^{1 \times 1, C \to C/r}(\boldsymbol{X})\big) \\
\boldsymbol{X}_l^{(2)} &= f_{\text{GELU}}\big(f_{\text{Conv}}^{3 \times 3}(\boldsymbol{X}_l^{(1)})\big) \\
\mathcal{F}_{LC}(\boldsymbol{X}) \equiv \boldsymbol{X}_l^{(\text{out})} &= f_{\text{Conv}}^{1 \times 1, C/r \to C}(\boldsymbol{X}_l^{(2)})
\end{aligned}
\end{equation}

\subsection{Effective Receptive Field (ERF) Derivation}
We provide a ERF analysis for the HLK branch, which is defined by the composition of the 1D horizontal convolutions ($1 \times k$). The ERF of a stack of convolutions is $\text{ERF}_{\text{out}} = \text{ERF}_{\text{in}} + (k - 1) \times d$, where $d$ is the dilation rate.

\paragraph{Standard Configuration ($\mathcal{F}_{LK\text{-S}}$).}
This configuration stacks $\mathcal{S}(\cdot, k_{core})$ and $\mathcal{S}_{d}(\cdot, k_{core}, d)$.
\begin{enumerate}
    \item The base $\mathcal{S}(\cdot, k_{core})$ block (specifically $f_{dw}^{1 \times k_{core}}$) establishes an $\text{ERF}_{\text{in}} = k_{core}$.
    \item The second stage, $\mathcal{S}_{d}(\cdot, k_{core}, d)$, (specifically $f_{dw}^{1 \times k_{core}, d}$) adds $(k_{core} - 1)d$.
\end{enumerate}
The total ERF is therefore:
\begin{equation}
    \text{ERF}_{\text{S}} = k_{core} + (k_{core} - 1)d \label{eq_erfs}
\end{equation}

\paragraph{Large Configuration ($\mathcal{F}_{LK\text{-L}}$).}
This configuration stacks $\mathcal{S}(\cdot, k_{core})$ and $\mathcal{S}_{d}(\cdot, k_{extra}, d)$.
\begin{enumerate}
    \item The base $\mathcal{S}(\cdot, k_{core})$ block establishes an $\text{ERF}_{\text{in}} = k_{core}$.
    \item The second stage, $\mathcal{S}_{d}(\cdot, k_{core}, d)$, adds $(k_{core} - 1)d$.
    \item The third stage, $\mathcal{S}_{d}(\cdot, k_{extra}, d)$, adds $(k_{extra} - 1)d$.
\end{enumerate}
The total ERF is therefore:
\begin{equation}
    \text{ERF}_{\text{L}} = k_{core} + (k_{core} - 1)d + (k_{extra} - 1)d \label{eq_erfl}
\end{equation}
This derivation confirms the formulas used to generate the configurations in Table~\ref{tab:lksa_specs}.

\begin{table}[ht!]
\centering
\caption{LKSA module configurations and resulting effective kernel sizes (1D ERF). For rows with $k_{\text{extra}} = \text{--}$, we use the $\mathcal{F}_{LK\text{-S}}$ formula \eqref{eq_erfs}; otherwise, we use the $\mathcal{F}_{LK\text{-L}}$ formula \eqref{eq_erfl}.}
\label{tab:lksa_specs}
\begin{tabular}{cccc}
\toprule
\textbf{$k_{\text{core}}$} & \textbf{Dilation $d$} & \textbf{$k_{\text{extra}}$} & \textbf{Final ERF} \\
\midrule
 3 & 1 & -- & 5  \\
 5 & 1 & -- & 9  \\
 \textbf{5} & \textbf{2} & -- & \textbf{13} \\
 5 & 3 & 11 & 47 \\
 5 & 3 & 13 & 53 \\
 5 & 3 & 17 & 65 \\
\bottomrule
\end{tabular}
\end{table}

\subsection{Feature Fusion and Final Output}
Finally, the outputs from the three branches are fused. The local and large-kernel spatial features are combined additively, and the result is modulated by the channel attention vector $\boldsymbol{X}_c$.
\begin{equation}
\begin{aligned}
    \boldsymbol{X}_{\text{spatial}} &= \mathcal{F}_{LC}(\boldsymbol{X}) + \mathcal{F}_{LK}(\boldsymbol{X}) \\
    \boldsymbol{X}_{\text{final}} &= \boldsymbol{X}_c \odot \boldsymbol{X}_{\text{spatial}}
\end{aligned}
\end{equation}
where $\odot$ denotes element-wise multiplication, with $\boldsymbol{X}_c$ being broadcast along the spatial dimensions. This allows $\boldsymbol{X}_l^{(\text{out})}$ to contribute fine-grained details, $\boldsymbol{X}_{lk}^{(\text{out})}$ to provide long-range context, and $\boldsymbol{X}_c$ to reweight the fused features based on channel-wise saliency.

\section{Limitations of ReLU and $\text{ELU} + 1$ in Linear Attention}
\label{subsec:relu_elu_limitations}
Recent linear attention architectures often adopt simple non-negative feature maps such as
ReLU or $\phi(x) = \operatorname{ELU}(x) + 1$.
However, we identify that both choices are suboptimal for single-image super-resolution,
albeit for different reasons.

\paragraph{ReLU feature map.}
The ReLU activation is defined as $\operatorname{ReLU}(x) = \max(0, x)$.
For a query--key pair $q_i, k_j \in \mathbb{R}^{d}$, the linearized attention kernel
becomes:
\begin{equation}
    \phi_{\text{ReLU}}(q_i)^{\top}\phi_{\text{ReLU}}(k_j)
    = \sum_{c=1}^{d} \max(0, q_{i,c}) \max(0, k_{j,c}).
\end{equation}
In contrast, standard softmax attention relies on the dot product
$q_i^{\top} k_j = \sum_{c} q_{i,c} k_{j,c}$,
where negative components are fully preserved:
if $q_{i,c} < 0$ and $k_{j,c} < 0$, their product is positive and contributes to similarity.
By forcing non-negativity, ReLU zeroes out these dimensions, discarding potentially informative
negative--negative alignments.
In super-resolution, where high-frequency details often rely on subtle, signed correlations
across channels, this information loss degrades the approximation of the softmax kernel.

\paragraph{$\text{ELU} + 1$ feature map.}
Let $\sigma(\cdot) = \operatorname{ELU}(\cdot)$. An alternative feature map is $\phi_{\text{ELU}+1}(x) = \sigma(x) + 1$.
For a single channel $c$, the interaction expands as:
\begin{equation}
    [\sigma(q_{i,c}) + 1][\sigma(k_{j,c}) + 1]
    = \sigma(q_{i,c})\sigma(k_{j,c}) + \sigma(q_{i,c}) + \sigma(k_{j,c}) + 1.
    \label{eq:elu_scalar}
\end{equation}
Summing over all $d$ channels yields the vector-form kernel:
\begin{equation}
\begin{aligned}
    \phi(q_i)^{\top} \phi(k_j)
    &= \sum_{c=1}^{d} \left[ \sigma(q_{i,c})\sigma(k_{j,c}) + \sigma(q_{i,c}) + \sigma(k_{j,c}) + 1 \right] \\
    &= \langle \sigma(q_i), \sigma(k_j) \rangle
     + \mathbf{1}^{\top}\sigma(q_i)
     + \mathbf{1}^{\top}\sigma(k_j)
     + d,
\end{aligned}
\label{eq:elu_vector}
\end{equation}
where $\mathbf{1} \in \mathbb{R}^{d}$ is the all-ones vector.
Crucially, the last three terms in Eq.~\eqref{eq:elu_vector} do not encode pairwise similarity;
they represent global query/key biases and a constant offset $d$.
In high-dimensional backbones, these bias terms (order $\mathcal{O}(d)$) often dominate the true similarity term $\langle \sigma(q_i), \sigma(k_j) \rangle$. This effectively compresses the attention variation into a narrow range, reducing the contrast between similar and dissimilar tokens and blurring high-frequency reconstruction details.

\begin{figure}[t]
    \centering
    \includegraphics[width=1\columnwidth]{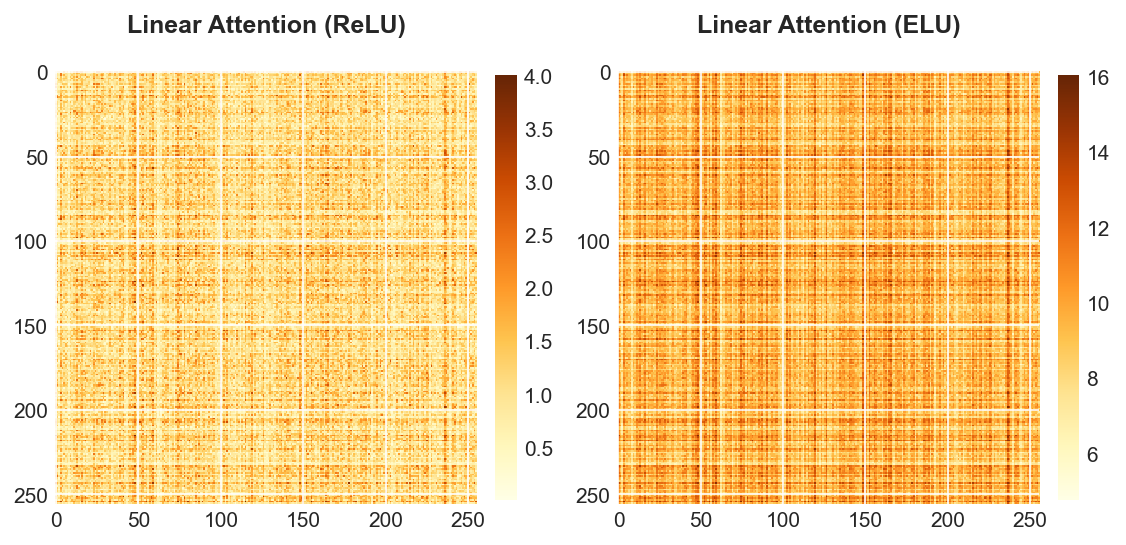}
    \caption{Visualization of attention maps for Linear Attention using ReLU and $\text{ELU} + 1$ (computed with sequence length $N=256$). The lack of normalization leads to high-magnitude artifacts.}
    \label{fig:elu_relu_atten_map}
    \vspace{-3mm}
\end{figure}

Figure~\ref{fig:elu_relu_atten_map} provides empirical validation of our analysis. We visualize the attention scores computed by the standard softmax operation (scaled by $1/\sqrt{d}$) versus the unnormalized linear kernels. Due to the lack of intrinsic normalization, both ReLU and $\text{ELU}+1$ produce attention scores
with significantly larger magnitudes. This effect is particularly pronounced for $\text{ELU}+1$, where scores surge to values
around $14$.
This empirical evidence corroborates the derivation above, confirming that the global bias terms in the $\text{ELU}+1$ kernel dominate the pairwise similarity signal.

\begin{figure}[t]
    \centering
    \includegraphics[width=1\columnwidth]{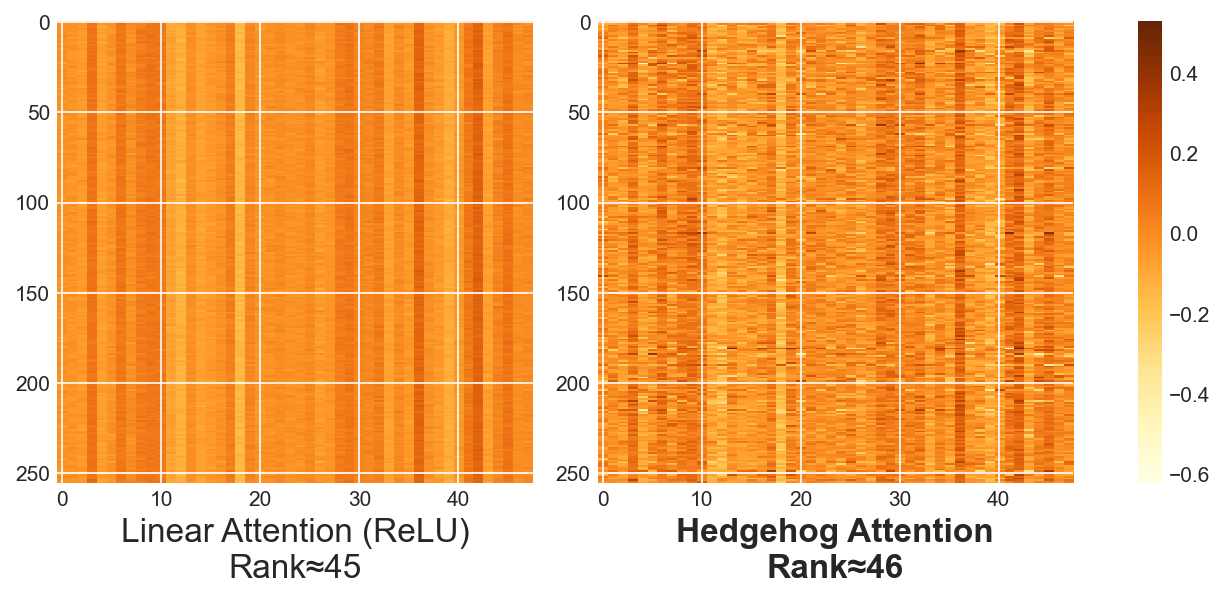}
    \caption{\textbf{Ranking consistency analysis.} We compare the output ranking of Linear Attention using standard ReLU, Symmetric ReLU, and the Hedgehog Feature Map (sequence length $N=256$). While adding negative information (Sym-ReLU) improves consistency, Hedgehog achieves superior performance through learnable stability.}
    \label{fig:relu_enhance}
    \vspace{-3mm}
\end{figure}

\paragraph{Ranking enhancement.}
To validate the hypothesis that discarding negative components degrades representational power,
we design a symmetric ReLU variant that explicitly preserves negative directionality via concatenation:
\begin{equation}
    \phi_{\text{sym}}(x) = [\operatorname{ReLU}(x), \operatorname{ReLU}(-x)].
\end{equation}
As illustrated in Fig.~\ref{fig:relu_enhance}, this symmetric formulation significantly improves
the output ranking consistency compared to the standard ReLU baseline.
However, this modification does not address the fundamental limitations of ReLU in linear attention,
namely its unbounded linear growth and lack of intrinsic normalization, which can lead to training instability.

Consequently, we adopt the Hedgehog feature map.
Unlike the fixed ReLU basis, Hedgehog employs a learnable Multi-Layer Perceptron (MLP) combined
with a Softmax activation.
This design offers two critical advantages:
(1) the Softmax ensures normalized, stable attention weights, and
(2) the MLP provides the flexibility to adaptively emphasize informative features across
both positive and negative regimes.
This analysis confirms that our architectural choice is principled rather than heuristic,
positioning our method as a robust, theoretically grounded alternative to recent Mamba-based \cite{guo2025mambair, guo2024mambairv2}
or standard Softmax \cite{park2025efficient} attention architectures.

\begin{table}[!b]
\centering
\caption{More ablation study. We train all models on DIV2K for 400K iterations, and test on Set5 and Urban100 ($\times2$).}
\label{tab:MoreAblationStudies}
\scalebox{0.75}{%
\renewcommand{\arraystretch}{1.2}
\begin{tabular}{llcccc}
\hline
\multirow{2}{*}{Case}                                                                       & \multirow{2}{*}{Configuration} & \multicolumn{2}{c}{Set5}                                                 & \multicolumn{2}{c}{Urban100}                                             \\ \cline{3-6} 
                                                                                            &                                & PSNR                               & SSIM                                & PSNR                               & SSIM                                \\ \hline
\multirow{3}{*}{Hybrid Block (HB)}                                                          & depth=3                        & 38.30                              & 0.9617                              & 33.05                              & 0.9363                              \\
                                                                                            & depth=2                        & 38.24                              & 0.9615                              & 32.95                              & 0.9353                              \\
                                                                                            & depth=1                        & 38.16                              & 0.9612                              & 32.71                              & 0.9336                              \\ \hline
\multirow{8}{*}{\begin{tabular}[c]{@{}l@{}}Large Kernel \\ Distillation (LKD)\end{tabular}} & LKD depth=4                    & 38.33                              & 0.9618                              & 33.12                              & 0.9369                              \\
                                                                                            & LKD depth=3                    & 38.33                              & 0.9618                              & 33.08                              & 0.9366                              \\
                                                                                            & LKD depth=2                    & 38.31                              & 0.9618                              & 33.04                              & 0.9363                              \\
                                                                                            & LKD depth=1                    & 38.31                              & 0.9617                              & 32.99                              & 0.9358                              \\ \cline{2-6} 
                                                                                            & Kernel size=5                  & 38.33                              & 0.9618                              & 33.12                              & 0.9369                              \\
                                                                                    
                                                                                            & Kernel size=53                 & 38.33                              & 0.9618                              & 33.18                              & 0.9375                              \\
                                                                                            & Kernel size=65                 & 38.33                              & 0.9618                              & 33.19                              & 0.9375                              \\ \hline
\textbf{UCAN}                                                                               & \textbf{Base}                  & \multicolumn{1}{l}{\textbf{38.34}} & \multicolumn{1}{l}{\textbf{0.9618}} & \multicolumn{1}{l}{\textbf{33.22}} & \multicolumn{1}{l}{\textbf{0.9379}} \\ \hline
\end{tabular}%
}
\end{table}
\section{More comparison with SOTA methods}

\begin{table*}[!ht]
\centering
\caption{Quantitative comparison on \textit{\textbf{lightweight image super-resolution}} with state-of-the-art methods. The best and second-best results are shown in \textbf{bold} and \underline{underlined}, respectively.}
\label{tab:lightSR}
\vspace{-3mm}
\setlength{\tabcolsep}{8pt}
\scalebox{0.70}{
\begin{tabular}{@{}l|c|c|c|cc|cc|cc|cc|cc@{}}
\toprule
 & & & & \multicolumn{2}{c|}{\textbf{Set5}} &
  \multicolumn{2}{c|}{\textbf{Set14}} &
  \multicolumn{2}{c|}{\textbf{BSDS100}} &
  \multicolumn{2}{c|}{\textbf{Urban100}} &
  \multicolumn{2}{c}{\textbf{Manga109}} \\
\multirow{-2}{*}{Method} & \multirow{-2}{*}{scale}& \multirow{-2}{*}{\#param}& \multirow{-2}{*}{FLOPs} & PSNR  & SSIM   & PSNR  & SSIM   & PSNR  & SSIM   & PSNR  & SSIM   & PSNR  & SSIM   \\ \midrule
% ---------------------- x2 ----------------------
SwinIR-light~\cite{liang2021swinir}& $2\times$  & 910K & {244.2G} 
& 38.14 & {0.9611} & {33.86} & {0.9206} & {32.31} & {0.9012} & {32.76} & {0.9340} & {39.12} & {0.9783} \\
ELAN~\cite{zhang2022efficient} & $2\times$  & 621K & 245.2G
& 38.27  &0.9616  &33.94 &0.9207  &32.30  &0.9012  &32.76  &0.9340 &39.11 &0.9782 \\
IPG-Tiny~\cite{tian2024image} & $2\times$  & 621K & 203.1G
& 38.17  &0.9611  &\underline{34.24} &\underline{0.9236}  &32.35  & 0.9018  &33.04  &0.9359 &39.31 &0.9786 \\
PFT-light~\cite{long2025progressive} & $2\times$  &776K &278.3G 
& \underline{38.36} &\textbf{0.9620} &34.19 &0.9232 &\underline{32.43} &\underline{0.9030} &\textbf{33.67} &\textbf{0.9411} &\underline{39.55} &\textbf{0.9792} \\
\textbf{UCAN (Our)} & $2\times$  &  689K & 146.3G 
& 38.34
& 0.9618
& \textbf{34.27}
& \textbf{0.9242}
& 32.39
& 0.9025	
& 33.22	
& 0.9379	
& 39.54
& \underline{0.9790}
\\
\textbf{UCAN-L (Our)}& $2\times$  &  886K & 182.4G 
& \textbf{38.37}
& \underline{0.9619}
& 34.19
& 0.9224
& \textbf{32.44}
& \textbf{0.9031}
& \underline{33.39}
& \underline{0.9393}
& \textbf{39.66}
& 0.9789
\\
\midrule
% ---------------------- x3 ----------------------
SwinIR-light~\cite{liang2021swinir} & $3\times$ & 918K & 111.2G 
& 34.62 & 0.9289 & 30.54 & 0.8463 & 29.20 & 0.8082 & 28.66 & 0.8624 & 33.98 & 0.9478 \\
ELAN~\cite{zhang2022efficient} & $3\times$ & 629K & 90.1G 
& 34.61 & 0.9288 & 30.55 & 0.8463 & 29.21 & 0.8081 & 28.69 & 0.8624 & 34.00 & 0.9478 \\
IPG-Tiny~\cite{tian2024image} & $3\times$  & 878K & 109.0G
& 34.64  &0.9292  &30.61 &0.8470  &29.26  & 0.8097  &28.93  & 0.8666 &34.30 &0.9493 \\
PFT-light~\cite{long2025progressive} & $3\times$ & 783K &123.5G 
& \underline{34.81} &0.9305 &\textbf{30.75} &\underline{0.8493} &\underline{29.33} &0.8116 &\textbf{29.43} &\textbf{0.8759} &\underline{34.60} &\underline{0.9510} \\
\textbf{UCAN (Our)} & $3\times$ & 696K & 64.6G 
& \textbf{34.83} & \underline{0.9308} & \underline{30.72} & \underline{0.8493} & \underline{29.32} & \underline{0.8121} & 29.15 & 0.8712 & \underline{34.62} & 0.9508 \\
\textbf{UCAN-L (Our)} & $3\times$ & 893K & 81.3G 
& \underline{34.81} & \textbf{0.9311} & \textbf{30.75} & \textbf{0.8500} & \textbf{29.34} & \textbf{0.8127} & \underline{29.29} & \underline{0.8738} & \textbf{34.79} & \textbf{0.9516} \\
\midrule
% ---------------------- x4 ----------------------
SwinIR-light~\cite{liang2021swinir} & $4\times$  & 930K & {63.6G} 
& {32.44}
& {0.8976}
& {28.77}
& {0.7858}
& {27.69}
& {0.7406}
& {26.47}
& {0.7980}
& {30.92}
& {0.9151}
\\
ELAN~\cite{zhang2022efficient} & $4\times$  & 640K& 54.1G 
&32.43 
&0.8975 
&28.78 
&0.7858 
&27.69 
&0.7406 
&26.54 
&0.7982 
&30.92
&0.9150
\\
IPG-Tiny~\cite{tian2024image} & $4\times$  & 887K & 61.3G 
&32.51 &0.8987 &28.85 &0.7873 &27.73& 0.7418 &26.78 &0.8050 &31.22 &0.9176 \\
PFT-light~\cite{long2025progressive} & $4\times$ & 792K & 69.6G 
& 32.63 & 0.9005&  28.92 & 0.7891 & 27.79 & 0.7445 & \textbf{27.20} & \textbf{0.8171} & \underline{31.51}&  \underline{0.9204} \\
\textbf{UCAN (Our)}& $4\times$  &  705K & 38.1G 
& \underline{32.65}
& \underline{0.9010}
& \underline{28.95}	
& \underline{0.7899}
& \underline{27.79}
& \underline{0.7454}	
& 26.89
& 0.8097
& 31.50	
& 0.9200
\\
\textbf{UCAN-L (Our)}& $4\times$  &  902K & 48.4G 
& \textbf{32.68}
& \textbf{0.9015}
& \textbf{28.99}	
& \textbf{0.7917}
& \textbf{27.80}
& \textbf{0.7459}	
& \underline{27.06}	
& \underline{0.8134}	
& \textbf{31.63}	
& \textbf{0.9212}
\\
\bottomrule
\end{tabular}%
}
\end{table*}

% Table~\ref{tab:lightSR} shows the comparison between our proposed model and other state-of-the-art lightweight models. The results demonstrate that our methods yield competitive performance. \\
% \paragraph{Computational Efficiency.} A key advantage of UCAN is its exceptional efficiency. At scale $\times 2$, UCAN requires only $146.3$G FLOPs, representing a reduction of approximately $\mathbf{47\%}$ compared to PFT-light ($278.3$G) and a significant decrease from SwinIR-light ($244.2$G). This efficiency advantage remains consistent across all scales. For instance, at scale $\times 3$, UCAN operates with the lowest computational cost ($64.6$G) among all compared methods. Similarly, at scale $\times 4$, it achieves state-of-the-art results using merely $38.1$G FLOPs—nearly half the computational cost of PFT-light ($69.6$G)—demonstrating a superior trade-off between model complexity and performance.

% \paragraph{Restoration Performance.} Despite the substantial reduction in computational cost, UCAN maintains highly competitive restoration quality. At scale $\times 4$, it achieves a PSNR of $32.65$ dB on Set5, surpassing both SwinIR-light ($32.44$ dB) and PFT-light ($32.63$ dB). Moreover, our scalable variant, UCAN-L, consistently secures the top or second-best performance across various benchmarks, further validating the robustness and flexibility of the proposed architecture.
Table~\ref{tab:lightSR} presents a comprehensive quantitative comparison with state-of-the-art lightweight methods. The results demonstrate that our proposed UCAN achieves a superior efficiency-performance trade-off by significantly reducing computational overhead while maintaining competitive restoration quality. Notably, at scale $\times 2$, UCAN requires only $146.3$G FLOPs. This represents a reduction of approximately $\mathbf{47\%}$ compared to PFT-light and a significant decrease relative to SwinIR-light. This efficiency advantage persists at scale $\times 4$ where UCAN operates with merely $38.1$G FLOPs, which is nearly half the computational cost of PFT-light. Despite this reduced complexity, UCAN delivers robust performance on challenging benchmarks such as Set14, BSDS100, and Manga109. For instance, on the Set14 dataset at scale $\times 2$, UCAN achieves a leading PSNR of $34.27$ dB. Moreover, our scalable variant UCAN-L consistently secures the best or second-best results across complex scenarios. On Manga109 at scale $\times 2$, UCAN-L surpasses the previous best method with a score of $39.66$ dB, further validating the effectiveness of our architecture in recovering fine structural details and textures.
\section{More ablation}

We investigate the contribution of each core component by systematically analyzing the depth of the Hybrid Blocks (HB), the Large Kernel Distillation (LKD) modules, and the choice of kernel size.
First, varying the \textbf{HB depth} from 3 to 1 reveals a steady performance decline; specifically, on Urban100, PSNR drops from $33.05$ dB to $32.71$ dB, confirming that multiple HB stages are essential for iteratively integrating local and global contexts.
A similar trend is observed when pruning the \textbf{LKD depth} from 4 to 1, where performance degrades from $33.12$ dB to $32.99$ dB, suggesting that deep distillation layers are critical for reconstructing intricate patterns.
Finally, we examine the impact of \textbf{kernel size} ($k \in \{5, 53, 65\}$) within the LKD. While performance on the simpler Set5 dataset saturates across all sizes ($38.33$ dB), larger kernels prove advantageous on the complex Urban100 benchmark, where increasing $k$ from 5 to 65 improves PSNR from $33.12$ dB to $33.19$ dB.
Based on these findings, our final UCAN Base model incorporates the optimal configuration, achieving a peak performance of $33.22$ dB.

\begin{figure*}[t]
    \centering
    \begin{minipage}{0.75\linewidth}
        \begin{tabular}{@{}c@{\hspace{4mm}}c@{}}
            % Ảnh gốc bên trái
            \begin{minipage}[c]{0.38\linewidth}
                \centering
                \includegraphics[width=\linewidth]{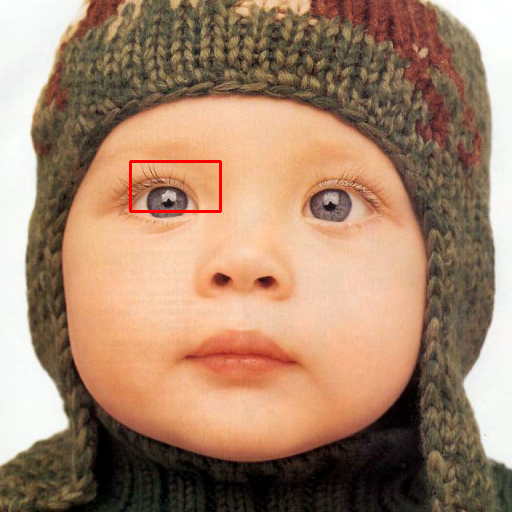} \\
                \vspace{1mm}
                {\small Set 5 - Baby}
            \end{minipage}
            &
            % Grid ảnh bên phải
            \begin{minipage}[c]{0.58\linewidth}
                \centering
                % Hàng 1
                \begin{minipage}[t]{0.31\linewidth}
                    \centering
                    \includegraphics[width=\linewidth]{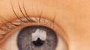} \\
                    {\tiny ATD \\ 27.18}
                \end{minipage}%
                \hfill
                \begin{minipage}[t]{0.31\linewidth}
                    \centering
                    \includegraphics[width=\linewidth]{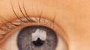} \\
                    {\tiny HiT-SRF \\ 27.35}
                \end{minipage}%
                \hfill
                \begin{minipage}[t]{0.31\linewidth}
                    \centering
                    \includegraphics[width=\linewidth]{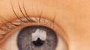} \\
                    {\tiny MambaIRv2 \\ 27.32}
                \end{minipage}

                \vspace{2mm}

                % Hàng 2
                \begin{minipage}[t]{0.31\linewidth}
                    \centering
                    \includegraphics[width=\linewidth]{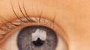} \\
                    {\tiny SRFormer \\ 27.13}
                \end{minipage}%
                \hfill
                \begin{minipage}[t]{0.31\linewidth}
                    \centering
                    \includegraphics[width=\linewidth]{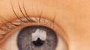} \\
                    {\tiny \textbf{UCAN} \\ 27.12}
                \end{minipage}%
                \hfill
                \begin{minipage}[t]{0.31\linewidth}
                    \centering
                    \includegraphics[width=\linewidth]{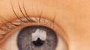} \\
                    {\tiny UCAN-L \\ \textbf{\textcolor{red}{29.37}}}
                \end{minipage}
            \end{minipage}
        \end{tabular}
    \end{minipage}
    \caption{Visual comparison between the ground truth and different methods on Set5 - baby.}
    \label{fig:visual_comparison_set5_baby}
\end{figure*}

\begin{figure*}[t]
    \centering
    \begin{minipage}{0.75\linewidth}
        \begin{tabular}{@{}c@{\hspace{4mm}}c@{}}
            % Ảnh gốc bên trái
            \begin{minipage}[c]{0.38\linewidth}
                \centering
                \includegraphics[width=\linewidth]{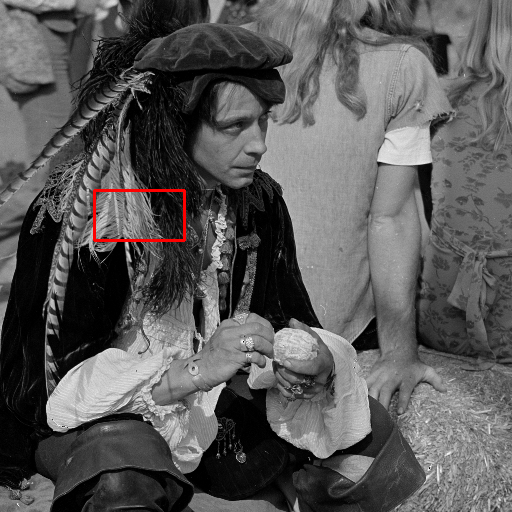} \\
                \vspace{1mm}
                {\small Set 14 - Man}
            \end{minipage}
            &
            % Grid ảnh bên phải
            \begin{minipage}[c]{0.58\linewidth}
                \centering
                % Hàng 1
                \begin{minipage}[t]{0.31\linewidth}
                    \centering
                    \includegraphics[width=\linewidth]{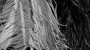} \\
                    {\tiny ATD \\ 20.42}
                \end{minipage}%
                \hfill
                \begin{minipage}[t]{0.31\linewidth}
                    \centering
                    \includegraphics[width=\linewidth]{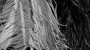} \\
                    {\tiny HiT-SRF \\ 20.47}
                \end{minipage}%
                \hfill
                \begin{minipage}[t]{0.31\linewidth}
                    \centering
                    \includegraphics[width=\linewidth]{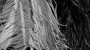} \\
                    {\tiny MambaIRv2 \\ 20.47}
                \end{minipage}

                \vspace{2mm}

                % Hàng 2
                \begin{minipage}[t]{0.31\linewidth}
                    \centering
                    \includegraphics[width=\linewidth]{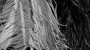} \\
                    {\tiny SRFormer \\ 20.55}
                \end{minipage}%
                \hfill
                \begin{minipage}[t]{0.31\linewidth}
                    \centering
                    \includegraphics[width=\linewidth]{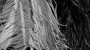} \\
                    {\tiny \textbf{UCAN} \\ 20.57}
                \end{minipage}%
                \hfill
                \begin{minipage}[t]{0.31\linewidth}
                    \centering
                    \includegraphics[width=\linewidth]{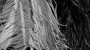} \\
                    {\tiny UCAN-L \\ \textbf{\textcolor{red}{20.59}}}
                \end{minipage}
            \end{minipage}
        \end{tabular}
    \end{minipage}
    \caption{Visual comparison between the ground truth and different methods on Set14 - man.}
    \label{fig:visual_comparison_set14_man}
\end{figure*}

\begin{figure*}[t]
    \centering
    \begin{minipage}{0.75\linewidth}
        \begin{tabular}{@{}c@{\hspace{4mm}}c@{}}
            % Ảnh gốc bên trái
            \begin{minipage}[c]{0.38\linewidth}
                \centering
                \includegraphics[width=\linewidth]{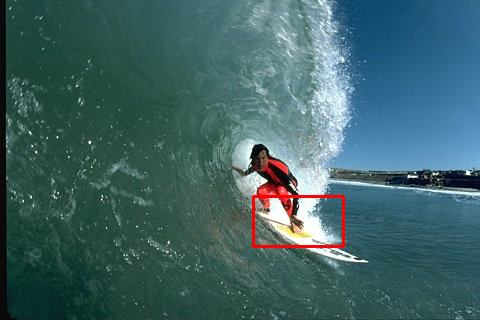} \\
                \vspace{1mm}
                {\small B100 - 300091}
            \end{minipage}
            &
            % Grid ảnh bên phải
            \begin{minipage}[c]{0.58\linewidth}
                \centering
                % Hàng 1
                \begin{minipage}[t]{0.31\linewidth}
                    \centering
                    \includegraphics[width=\linewidth]{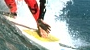} \\
                    {\tiny ATD \\ 22.28}
                \end{minipage}%
                \hfill
                \begin{minipage}[t]{0.31\linewidth}
                    \centering
                    \includegraphics[width=\linewidth]{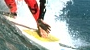} \\
                    {\tiny HiT-SRF \\ 22.33}
                \end{minipage}%
                \hfill
                \begin{minipage}[t]{0.31\linewidth}
                    \centering
                    \includegraphics[width=\linewidth]{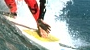} \\
                    {\tiny MambaIRv2 \\ 22.25}
                \end{minipage}

                \vspace{2mm}

                % Hàng 2
                \begin{minipage}[t]{0.31\linewidth}
                    \centering
                    \includegraphics[width=\linewidth]{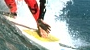} \\
                    {\tiny SRFormer \\ 22.19}
                \end{minipage}%
                \hfill
                \begin{minipage}[t]{0.31\linewidth}
                    \centering
                    \includegraphics[width=\linewidth]{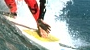} \\
                    {\tiny \textbf{UCAN} \\ 22.34}
                \end{minipage}%
                \hfill
                \begin{minipage}[t]{0.31\linewidth}
                    \centering
                    \includegraphics[width=\linewidth]{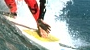} \\
                    {\tiny UCAN-L \\ \textbf{\textcolor{red}{22.37}}}
                \end{minipage}
            \end{minipage}
        \end{tabular}
    \end{minipage}
    \caption{Visual comparison between the ground truth and different methods on B100 - 300091.}
    \label{fig:visual_comparison_b100_300091}
\end{figure*}

\begin{figure*}[t]
    \centering
    \begin{minipage}{0.75\linewidth}
        \begin{tabular}{@{}c@{\hspace{4mm}}c@{}}
            % Ảnh gốc bên trái
            \begin{minipage}[c]{0.38\linewidth}
                \centering
                \includegraphics[width=\linewidth]{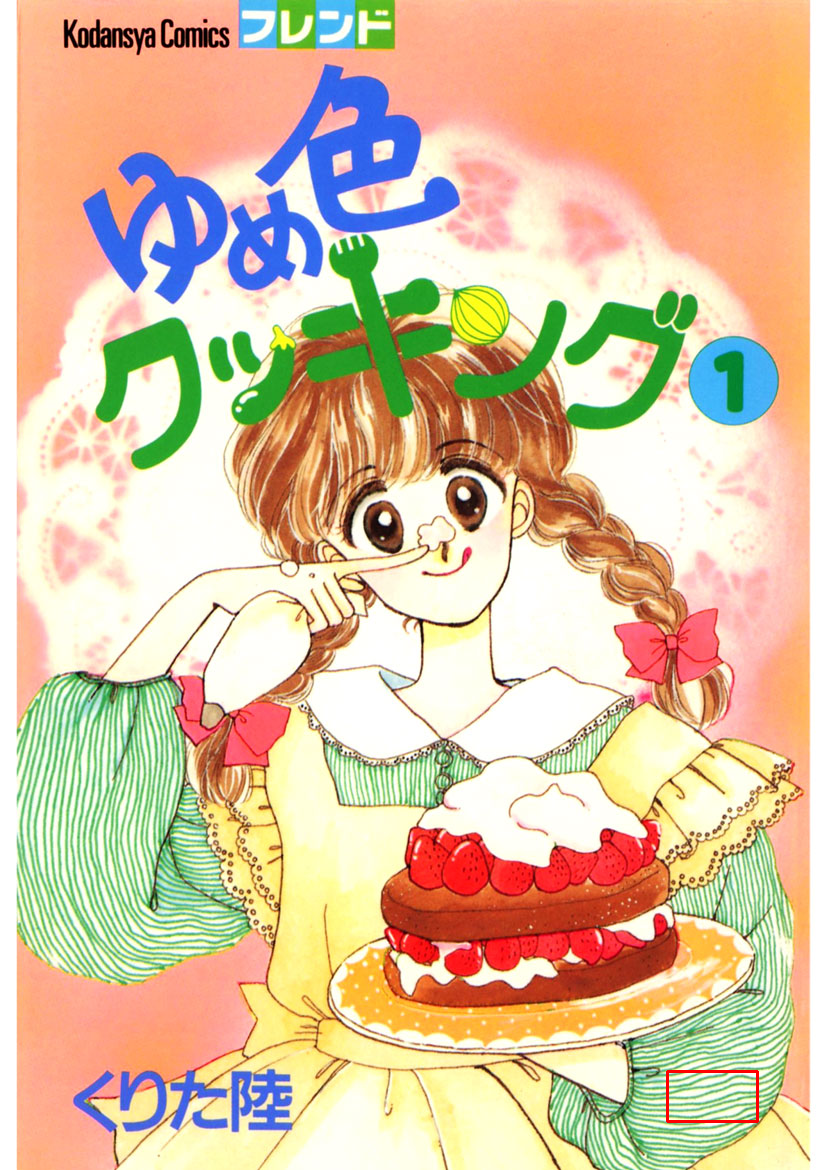} \\
                \vspace{1mm}
                {\small Manga109 - Yumeko Cooking}
            \end{minipage}
            &
            % Grid ảnh bên phải
            \begin{minipage}[c]{0.58\linewidth}
                \centering
                % Hàng 1
                \begin{minipage}[t]{0.31\linewidth}
                    \centering
                    \includegraphics[width=\linewidth]{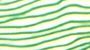} \\
                    {\tiny ATD \\ 23.55}
                \end{minipage}%
                \hfill
                \begin{minipage}[t]{0.31\linewidth}
                    \centering
                    \includegraphics[width=\linewidth]{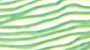} \\
                    {\tiny HiT-SRF \\ 22.68}
                \end{minipage}%
                \hfill
                \begin{minipage}[t]{0.31\linewidth}
                    \centering
                    \includegraphics[width=\linewidth]{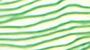} \\
                    {\tiny MambaIRv2 \\24.55}
                \end{minipage}

                \vspace{2mm}

                % Hàng 2
                \begin{minipage}[t]{0.31\linewidth}
                    \centering
                    \includegraphics[width=\linewidth]{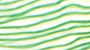} \\
                    {\tiny SRFormer \\ 22.17}
                \end{minipage}%
                \hfill
                \begin{minipage}[t]{0.31\linewidth}
                    \centering
                    \includegraphics[width=\linewidth]{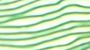} \\
                    {\tiny \textbf{UCAN} \\ 23.23}
                \end{minipage}%
                \hfill
                \begin{minipage}[t]{0.31\linewidth}
                    \centering
                    \includegraphics[width=\linewidth]{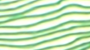} \\
                    {\tiny UCAN-L \\ \textbf{\textcolor{red}{24.92}}}
                \end{minipage}
            \end{minipage}
        \end{tabular}
    \end{minipage}
    \caption{Visual comparison between the ground truth and different methods on Manga109 - Yumeko Cooking.}
    \label{fig:visual_comparison_manga109_cooking}
    \end{figure*}
\begin{figure*}[t]
    \centering
    \begin{minipage}{0.85\linewidth}
        \begin{tabular}{@{}c@{\hspace{4mm}}c@{}}
            % Ảnh gốc bên trái
            \begin{minipage}[c]{0.38\linewidth}
                \centering
                \includegraphics[width=\linewidth]{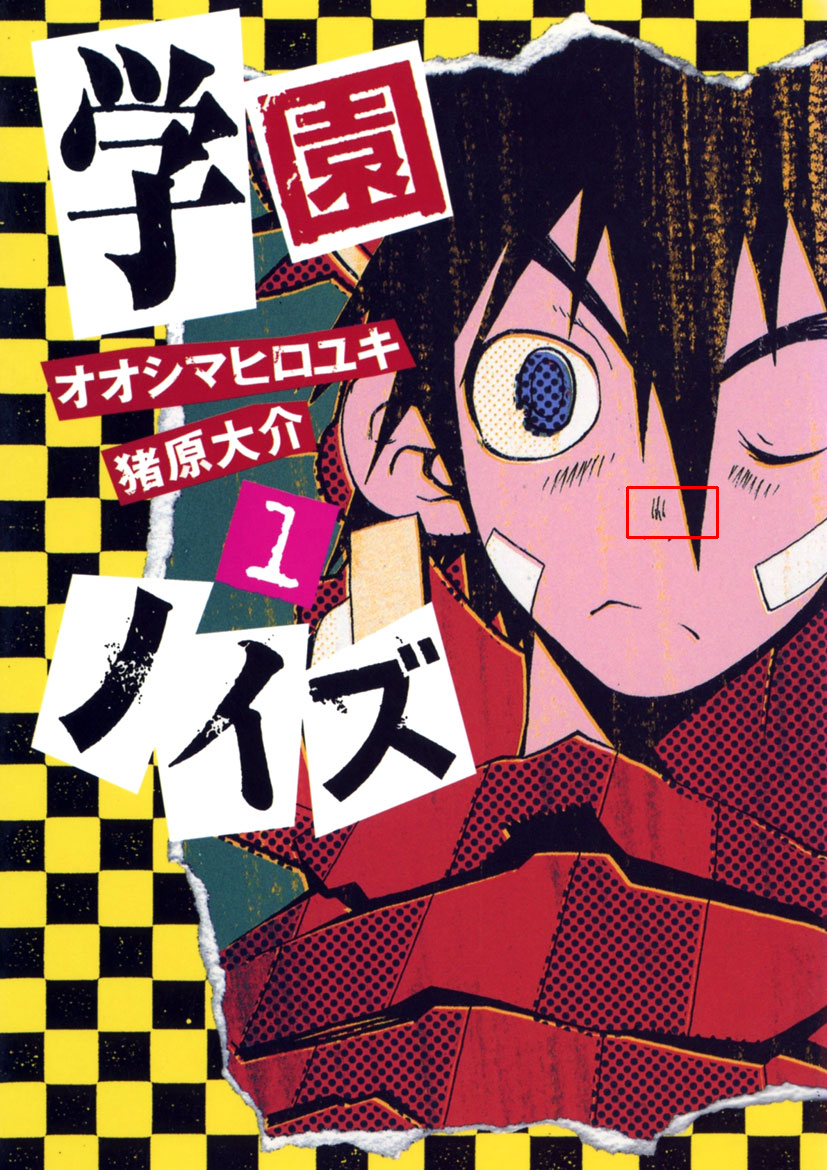} \\
                \vspace{1mm}
                {\small Manga109 - Gakuen Noise}
            \end{minipage}
            &
            % Grid ảnh bên phải
            \begin{minipage}[c]{0.58\linewidth}
                \centering
                % Hàng 1
                \begin{minipage}[t]{0.31\linewidth}
                    \centering
                    \includegraphics[width=\linewidth]{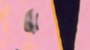} \\
                    {\tiny ATD \\ 27.34}
                \end{minipage}%
                \hfill
                \begin{minipage}[t]{0.31\linewidth}
                    \centering
                    \includegraphics[width=\linewidth]{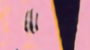} \\
                    {\tiny HiT-SRF \\ 27.72}
                \end{minipage}%
                \hfill
                \begin{minipage}[t]{0.31\linewidth}
                    \centering
                    \includegraphics[width=\linewidth]{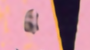} \\
                    {\tiny MambaIRv2 \\ 27.11}
                \end{minipage}

                \vspace{2mm}

                % Hàng 2
                \begin{minipage}[t]{0.31\linewidth}
                    \centering
                    \includegraphics[width=\linewidth]{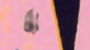} \\
                    {\tiny SRFormer \\ 26.74}
                \end{minipage}%
                \hfill
                \begin{minipage}[t]{0.31\linewidth}
                    \centering
                    \includegraphics[width=\linewidth]{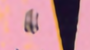} \\
                    {\tiny \textbf{UCAN} \\ 27.57}
                \end{minipage}%
                \hfill
                \begin{minipage}[t]{0.31\linewidth}
                    \centering
                    \includegraphics[width=\linewidth]{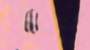} \\
                    {\tiny UCAN-L \\ \textbf{\textcolor{red}{28.10}}}
                \end{minipage}
            \end{minipage}
        \end{tabular}
    \end{minipage}
    \caption{Visual comparison between the ground truth and different methods on Manga109 - Gakuen Noise.}
    \label{fig:visual_comparison_manga109_gakuen}
\end{figure*}

\begin{figure*}[t]
    \centering
    \begin{minipage}{0.85\linewidth}
        \begin{tabular}{@{}c@{\hspace{4mm}}c@{}}
            % Ảnh gốc bên trái
            \begin{minipage}[c]{0.38\linewidth}
                \centering
                \includegraphics[width=\linewidth]{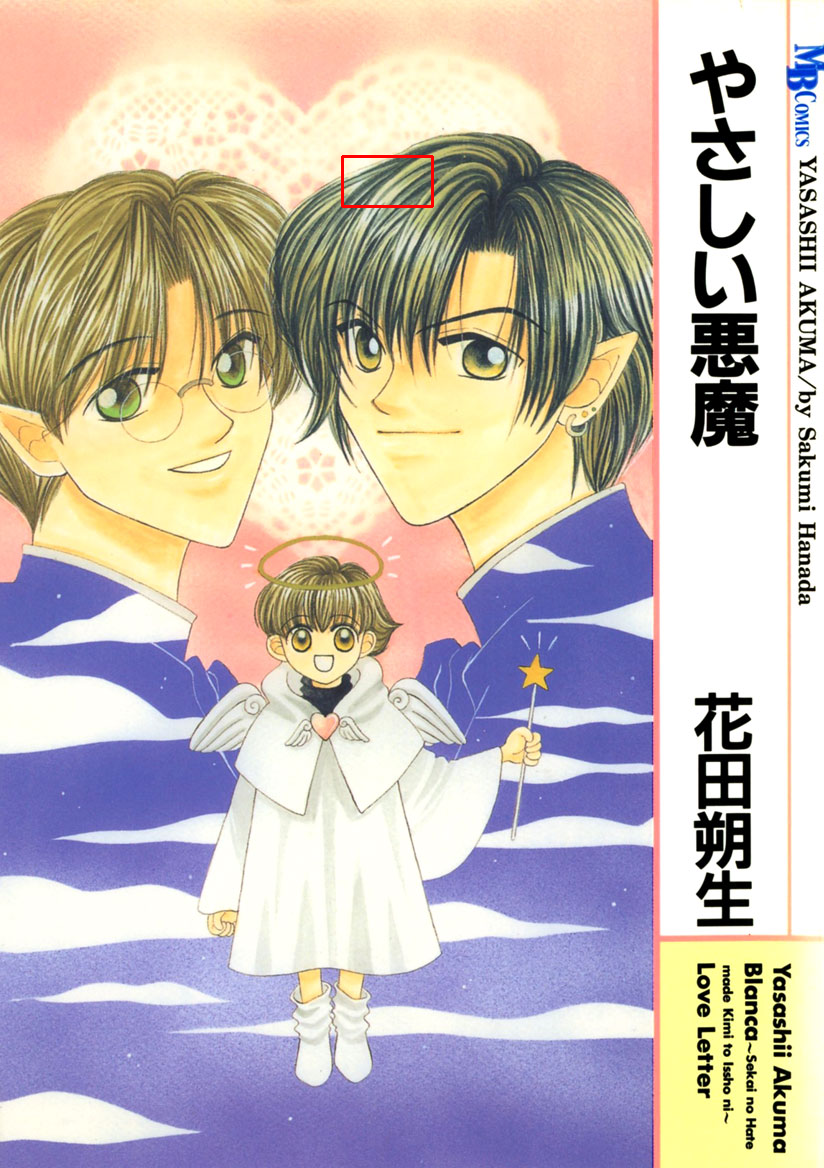} \\
                \vspace{1mm}
                {\small Manga109 - Yasakii Akuma}
            \end{minipage}
            &
            % Grid ảnh bên phải
            \begin{minipage}[c]{0.58\linewidth}
                \centering
                % Hàng 1
                \begin{minipage}[t]{0.31\linewidth}
                    \centering
                    \includegraphics[width=\linewidth]{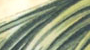} \\
                    {\tiny ATD \\ 25.68}
                \end{minipage}%
                \hfill
                \begin{minipage}[t]{0.31\linewidth}
                    \centering
                    \includegraphics[width=\linewidth]{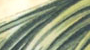} \\
                    {\tiny HiT-SRF \\ 25.79}
                \end{minipage}%
                \hfill
                \begin{minipage}[t]{0.31\linewidth}
                    \centering
                    \includegraphics[width=\linewidth]{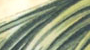} \\
                    {\tiny MambaIRv2 \\ 25.41}
                \end{minipage}

                \vspace{2mm}

                % Hàng 2
                \begin{minipage}[t]{0.31\linewidth}
                    \centering
                    \includegraphics[width=\linewidth]{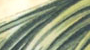} \\
                    {\tiny SRFormer \\ 25.06}
                \end{minipage}%
                \hfill
                \begin{minipage}[t]{0.31\linewidth}
                    \centering
                    \includegraphics[width=\linewidth]{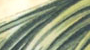} \\
                    {\tiny \textbf{UCAN} \\ 25.58}
                \end{minipage}%
                \hfill
                \begin{minipage}[t]{0.31\linewidth}
                    \centering
                    \includegraphics[width=\linewidth]{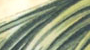} \\
                    {\tiny UCAN-L \\ \textbf{\textcolor{red}{26.59}}}
                \end{minipage}
            \end{minipage}
        \end{tabular}
    \end{minipage}
    \caption{Visual comparison between the ground truth and different methods on Manga109 - Yasakii Akuma.}
    \label{fig:visual_comparison_manga109_yasakii}
\end{figure*}

\begin{figure*}[t]
    \centering
    \begin{minipage}{0.85\linewidth}
        \begin{tabular}{@{}c@{\hspace{4mm}}c@{}}
            % Ảnh gốc bên trái
            \begin{minipage}[c]{0.38\linewidth}
                \centering
                \includegraphics[width=\linewidth]{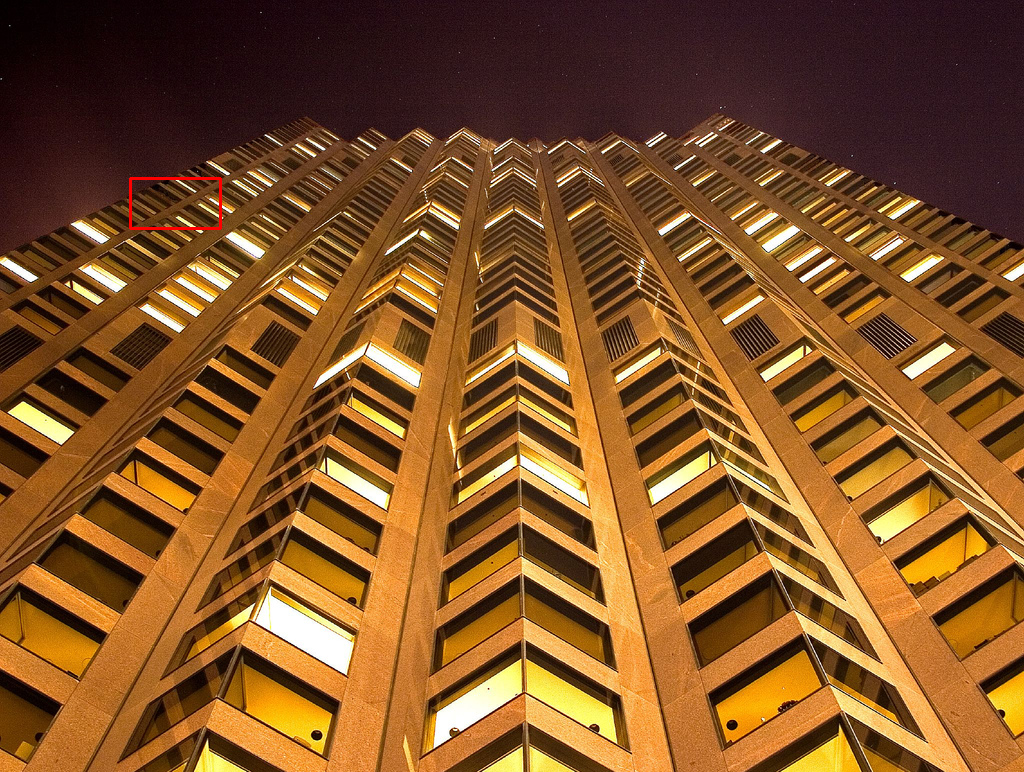} \\
                \vspace{1mm}
                {\small Urban100 - img15}
            \end{minipage}
            &
            % Grid ảnh bên phải
            \begin{minipage}[c]{0.58\linewidth}
                \centering
                % Hàng 1
                \begin{minipage}[t]{0.31\linewidth}
                    \centering
                    \includegraphics[width=\linewidth]{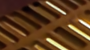} \\
                    {\tiny ATD \\ 24.58}
                \end{minipage}%
                \hfill
                \begin{minipage}[t]{0.31\linewidth}
                    \centering
                    \includegraphics[width=\linewidth]{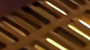} \\
                    {\tiny HiT-SRF \\ 25.02}
                \end{minipage}%
                \hfill
                \begin{minipage}[t]{0.31\linewidth}
                    \centering
                    \includegraphics[width=\linewidth]{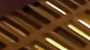} \\
                    {\tiny MambaIRv2 \\ 24.93}
                \end{minipage}

                \vspace{2mm}

                % Hàng 2
                \begin{minipage}[t]{0.31\linewidth}
                    \centering
                    \includegraphics[width=\linewidth]{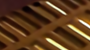} \\
                    {\tiny SRFormer \\ 24.75}
                \end{minipage}%
                \hfill
                \begin{minipage}[t]{0.31\linewidth}
                    \centering
                    \includegraphics[width=\linewidth]{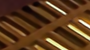} \\
                    {\tiny \textbf{UCAN} \\ 25.21}
                \end{minipage}%
                \hfill
                \begin{minipage}[t]{0.31\linewidth}
                    \centering
                    \includegraphics[width=\linewidth]{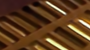} \\
                    {\tiny UCAN-L \\ \textbf{\textcolor{red}{25.31}}}
                \end{minipage}
            \end{minipage}
        \end{tabular}
    \end{minipage}
    \caption{Visual comparison between the ground truth and different methods on Urban100 - 015.}
    \label{fig:visual_comparison_urban100_015}
\end{figure*}

\begin{figure*}[t]
    \centering
    \begin{minipage}{0.85\linewidth}
        \begin{tabular}{@{}c@{\hspace{4mm}}c@{}}
            % Ảnh gốc bên trái
            \begin{minipage}[c]{0.38\linewidth}
                \centering
                \includegraphics[width=\linewidth]{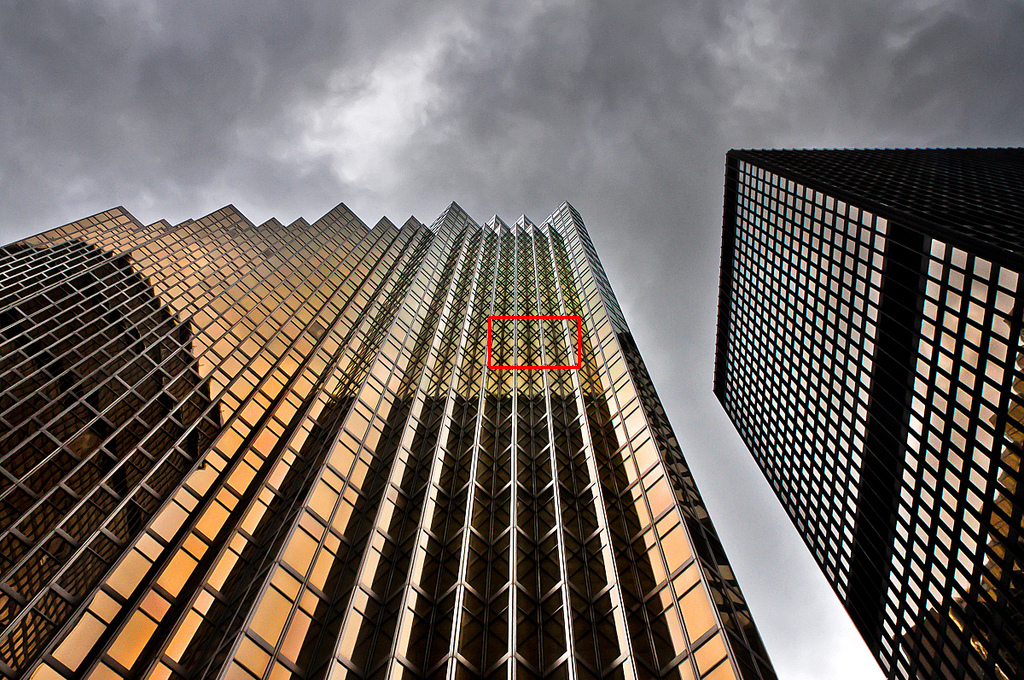} \\
                \vspace{1mm}
                {\small Urban100 - img19}
            \end{minipage}
            &
            % Grid ảnh bên phải
            \begin{minipage}[c]{0.58\linewidth}
                \centering
                % Hàng 1
                \begin{minipage}[t]{0.31\linewidth}
                    \centering
                    \includegraphics[width=\linewidth]{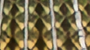} \\
                    {\tiny ATD \\ 13.80}
                \end{minipage}%
                \hfill
                \begin{minipage}[t]{0.31\linewidth}
                    \centering
                    \includegraphics[width=\linewidth]{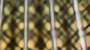} \\
                    {\tiny HiT-SRF \\ 13.77}
                \end{minipage}%
                \hfill
                \begin{minipage}[t]{0.31\linewidth}
                    \centering
                    \includegraphics[width=\linewidth]{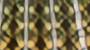} \\
                    {\tiny MambaIRv2 \\ 13.79}
                \end{minipage}

                \vspace{2mm}

                % Hàng 2
                \begin{minipage}[t]{0.31\linewidth}
                    \centering
                    \includegraphics[width=\linewidth]{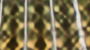} \\
                    {\tiny SRFormer \\ 13.48}
                \end{minipage}%
                \hfill
                \begin{minipage}[t]{0.31\linewidth}
                    \centering
                    \includegraphics[width=\linewidth]{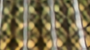} \\
                    {\tiny \textbf{UCAN} \\ 13.82}
                \end{minipage}%
                \hfill
                \begin{minipage}[t]{0.31\linewidth}
                    \centering
                    \includegraphics[width=\linewidth]{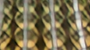} \\
                    {\tiny UCAN-L \\ \textbf{\textcolor{red}{13.85}}}
                \end{minipage}
            \end{minipage}
        \end{tabular}
    \end{minipage}
    \caption{Visual comparison between the ground truth and different methods on Urban100 - img19.}
    \label{fig:visual_comparison_urban100_019}
\end{figure*}

\begin{figure*}[t]
    \centering
    \begin{minipage}{0.85\linewidth}
        \begin{tabular}{@{}c@{\hspace{4mm}}c@{}}
            % Ảnh gốc bên trái
            \begin{minipage}[c]{0.38\linewidth}
                \centering
                \includegraphics[width=\linewidth]{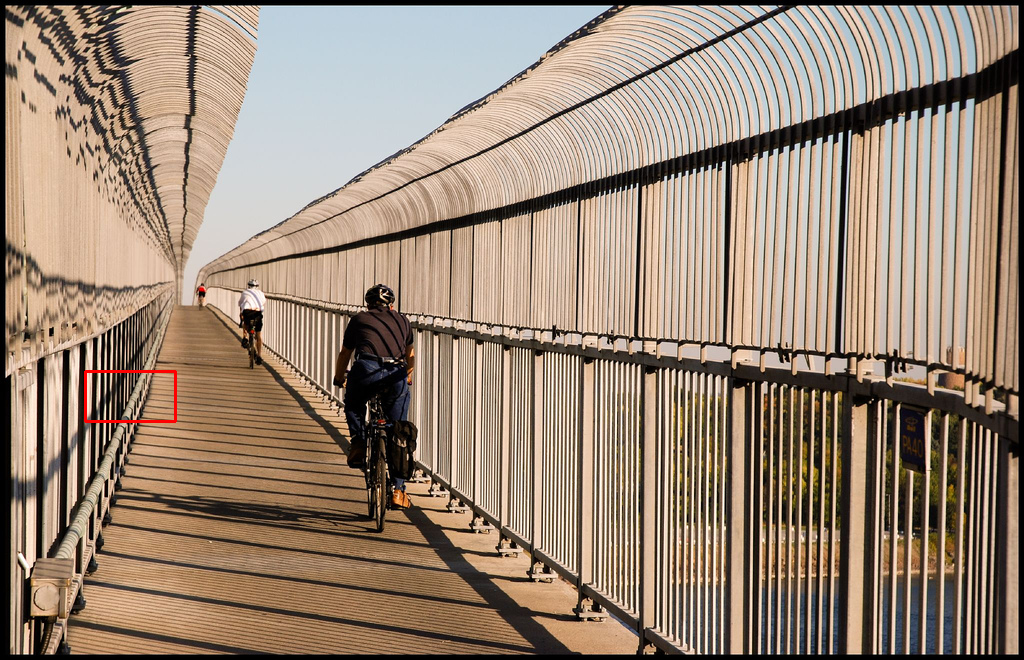} \\
                \vspace{1mm}
                {\small Urban100 - img24}
            \end{minipage}
            &
            % Grid ảnh bên phải
            \begin{minipage}[c]{0.58\linewidth}
                \centering
                % Hàng 1
                \begin{minipage}[t]{0.31\linewidth}
                    \centering
                    \includegraphics[width=\linewidth]{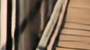} \\
                    {\tiny ATD \\ 15.63}
                \end{minipage}%
                \hfill
                \begin{minipage}[t]{0.31\linewidth}
                    \centering
                    \includegraphics[width=\linewidth]{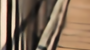} \\
                    {\tiny HiT-SRF \\ 16.64}
                \end{minipage}%
                \hfill
                \begin{minipage}[t]{0.31\linewidth}
                    \centering
                    \includegraphics[width=\linewidth]{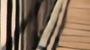} \\
                    {\tiny MambaIRv2 \\ 17.07}
                \end{minipage}

                \vspace{2mm}

                % Hàng 2
                \begin{minipage}[t]{0.31\linewidth}
                    \centering
                    \includegraphics[width=\linewidth]{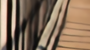} \\
                    {\tiny SRFormer \\ 16.29}
                \end{minipage}%
                \hfill
                \begin{minipage}[t]{0.31\linewidth}
                    \centering
                    \includegraphics[width=\linewidth]{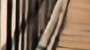} \\
                    {\tiny \textbf{UCAN} \\ \textbf{\textcolor{red}{17.74}}}
                \end{minipage}%
                \hfill
                \begin{minipage}[t]{0.31\linewidth}
                    \centering
                    \includegraphics[width=\linewidth]{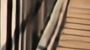} \\
                    {\tiny UCAN-L \\ 16.94}
                \end{minipage}
            \end{minipage}
        \end{tabular}
    \end{minipage}
    \caption{Visual comparison between the ground truth and different methods on Urban100 - img24.}
    \label{fig:visual_comparison_urban100_024}
\end{figure*}

\begin{figure*}[t]
    \centering
    \begin{minipage}{0.85\linewidth}
        \begin{tabular}{@{}c@{\hspace{4mm}}c@{}}
            % Ảnh gốc bên trái
            \begin{minipage}[c]{0.38\linewidth}
                \centering
                \includegraphics[width=\linewidth]{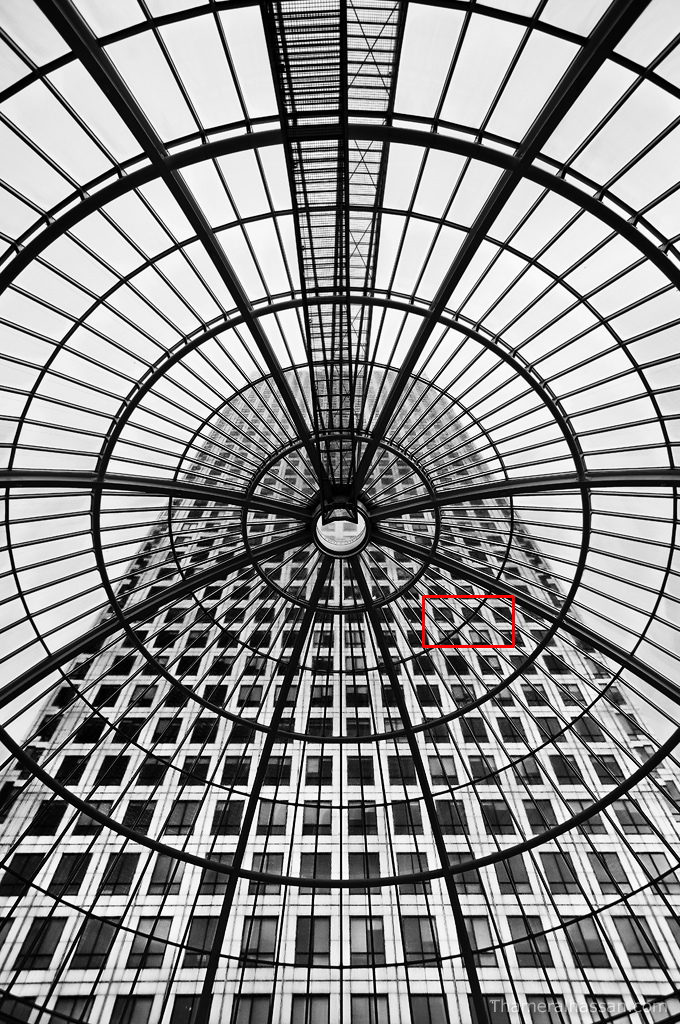} \\
                \vspace{1mm}
                {\small Urban100 - img72}
            \end{minipage}
            &
            % Grid ảnh bên phải
            \begin{minipage}[c]{0.58\linewidth}
                \centering
                % Hàng 1
                \begin{minipage}[t]{0.31\linewidth}
                    \centering
                    \includegraphics[width=\linewidth]{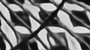} \\
                    {\tiny ATD \\ 14.69}
                \end{minipage}%
                \hfill
                \begin{minipage}[t]{0.31\linewidth}
                    \centering
                    \includegraphics[width=\linewidth]{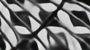} \\
                    {\tiny HiT-SRF \\ 14.63}
                \end{minipage}%
                \hfill
                \begin{minipage}[t]{0.31\linewidth}
                    \centering
                    \includegraphics[width=\linewidth]{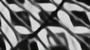} \\
                    {\tiny MambaIRv2 \\ 14.58}
                \end{minipage}

                \vspace{2mm}

                % Hàng 2
                \begin{minipage}[t]{0.31\linewidth}
                    \centering
                    \includegraphics[width=\linewidth]{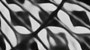} \\
                    {\tiny SRFormer \\ 14.71}
                \end{minipage}%
                \hfill
                \begin{minipage}[t]{0.31\linewidth}
                    \centering
                    \includegraphics[width=\linewidth]{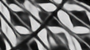} \\
                    {\tiny \textbf{UCAN} \\ 14.60 }
                \end{minipage}%
                \hfill
                \begin{minipage}[t]{0.31\linewidth}
                    \centering
                    \includegraphics[width=\linewidth]{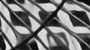} \\
                   {\tiny UCAN-L \\ \textbf{\textcolor{red}{14.93}}}

                \end{minipage}
            \end{minipage}
        \end{tabular}
    \end{minipage}
    \caption{Visual comparison between the ground truth and different methods on Urban100 - img72.}
    \label{fig:visual_comparison_urban100_072}
\end{figure*}

\end{document}